\documentclass{article}
\usepackage{arxiv}
\usepackage{cite}
\usepackage{amsmath,amssymb,amsfonts}
\usepackage{algorithmic}
\usepackage{graphicx}
\usepackage{soul}
\usepackage{textcomp}
\usepackage{comment}
\def\BibTeX{{\rm B\kern-.05em{\sc i\kern-.025em b}\kern-.08em
    T\kern-.1667em\lower.7ex\hbox{E}\kern-.125emX}}

\usepackage{cite}
\usepackage{qcircuit}
\usepackage{caption}
\usepackage{subcaption}
\usepackage{amsmath}
\usepackage{algorithmic}
\usepackage{tabularx}
\usepackage{xcolor}
\usepackage{tabularray}
\UseTblrLibrary{booktabs}
\newcolumntype{?}{!{\vrule width 1.5pt}}
\usepackage{colortbl}
\definecolor{LightGreen}{rgb}{0.812,1,0.812}
\definecolor{LightBlue}{rgb}{0.706,0.95,1}
\definecolor{LightOrange}{rgb}{1, 0.9254, 0.86}
\definecolor{LightYellow}{rgb}{1, 0.976, 0.812}
\usepackage{url}
\usepackage{hyperref}
\usepackage{array}
\usepackage{braket}
\usepackage{float}
\usepackage{ulem}
\newenvironment{conditions*}
  {\par\vspace{\abovedisplayskip}\noindent
   \tabularx{\columnwidth}{>{$}l<{$} @{${}={}$} >{\raggedright\arraybackslash}X}}
  {\endtabularx\par\vspace{\belowdisplayskip}}

\usepackage[abbreviations]{glossaries-extra}

\glssetcategoryattribute{acronym}{indexonlyfirst}{true}

\newabbreviation{qc}{QC}{Quantum Computing}
\newabbreviation{qml}{QML}{Quantum Machine Learning}
\newabbreviation{ml}{ML}{Machine Learning}
\newabbreviation{vqa}{VQA}{Variational Quantum Algorithm}
\newabbreviation{vqc}{VQC}{Variational Quantum Circuit}

\begin{document}
\sloppy

\pagestyle{fancy}

\title{Evaluating Angle and Amplitude Encoding Strategies for Variational Quantum Machine Learning: their impact on model's accuracy}

\author{
  Antonio Tudisco \\
  Department of Electronics and Telecommunications\\
  Politecnico di Torino, Turin 10129, Italy\\
  \texttt{antonio.tudisco@polito.it}
\And
Andrea Marchesin \\
  Department of Electronics and Nanoengineering\\
  Aalto University, 02150 Espoo, Finland\\
  \texttt{andrea.marchesin@aalto.fi}
\And
  Maurizio Zamboni \\
  Department of Electronics and Telecommunications\\
  Politecnico di Torino, Turin 10129, Italy\\
  \texttt{maurizio.zamboni@polito.it}
\And
  Mariagrazia Graziano \\
  Department of Applied Science and Technology\\
  Politecnico di Torino, Turin 10129, Italy\\
  \texttt{mariagrazia.graziano@polito.it}
  \And
  Giovanna Turvani \\
  Department of Electronics and Telecommunications\\
  Politecnico di Torino, Turin 10129, Italy\\
  \texttt{giovanna.turvani@polito.it}
}

\maketitle


\begin{abstract}
Recent advancements in Quantum Computing and Machine Learning have increased attention to Quantum Machine Learning (QML), which aims to develop machine learning models by exploiting the quantum computing paradigm. One of the widely used models in this area is the Variational Quantum Circuit (VQC), a hybrid model where the quantum circuit handles data inference while classical optimization adjusts the parameters of the circuit. The quantum circuit consists of an encoding layer, which loads data into the circuit, and a template circuit, known as the ansatz, responsible for processing the data.
This work involves performing an analysis by considering both Amplitude- and Angle-encoding models, and examining how the type of rotational gate applied affects the classification performance of the model. This comparison is carried out by training the different models on two datasets, Wine and Diabetes, and evaluating their performance.
The study demonstrates that, under identical model topologies, the difference in accuracy between the best and worst models ranges from 10\% to 30\%, with differences reaching up to 41\%. Moreover, the results highlight how the choice of rotational gates used in encoding can significantly impact the model's classification performance. The findings confirm that the embedding represents a hyperparameter for VQC models.
\end{abstract}

\keywords{Quantum Embedding, Quantum Machine Learning, Variational Quantum Algorithms.}

\section{Introduction}
In the contemporary era, information technology has become ubiquitous with profound implications for society. The advent of new electronic devices and connected services has enabled the collection of significant amounts of data, which can be further leveraged to improve the quality of our daily lives. However, interpreting data from various sources is a complex task, and the extraction of useful information requires sophisticated algorithms and technologies. Artificial Intelligence (AI), specifically within the branch of \gls*{ml}\cite{ML_bible}, offers a promising solution to automate this elaboration process. \\
Over the past three decades, many algorithms have been developed to accomplish precise tasks, with widespread applications in technology and science domains, including autonomous vehicle control, robotics, natural language processing, and computer vision\cite{ML_rev}. However, these solutions are computationally expensive, and there is a need to find new efficient strategies to perform the related tasks.\\

In parallel with AI research, \gls*{qc}\cite{QC_bible, QSim_Lloyd} has gained significant attention from the scientific community due to its potential to solve complex problems with algorithms that theoretically prove superior to classical known approaches (e.g., Shor's factorization algorithm\cite{Shor} and Grover's quantum search algorithm \cite{10.1145/237814.237866}). This potential, coupled with recent technological advancements that have led to the development of the first quantum computers, has prompted research to investigate opportunities for developing new \gls*{qc} solutions to address real-world complex problems. \\
The objective is to expand the computational capabilities of modern processing systems beyond limits previously considered unimaginable. Today, researchers are seeking to initiate another revolution based on this new paradigm, similar to the introduction of the first computers in the 1970s, to find applications with a quantum advantage.\\

Given the computational requirements of current \gls*{ml} algorithms and the prospect of a quantum advantage in the field, \gls*{qml}\cite{Schuld_first, Biamonte_wittek, Ciliberto, Schuld-basics_1, Schuld-basics-2} was born. Several models have been proposed in the last few years to address a variety of tasks, ranging from solving high-energy physics problems \cite{QML_HEP}helping with the detection of drowsiness through EEG signals \cite{QML_drowsiness}, to improving applications in chemical engineering such as reaction kinetics and optimization \cite{Bernal_QC_chem_eng}, to discovering new drugs \cite{QML_drugs}, defining new convolutional neural networks \cite{QCNN}, and even UAV path planning using a quantum‑inspired experience replay in a Deep Reinforcement Learning framework \cite{9748970}. The research in this new field is still in its infancy with models demonstrating limited capabilities; thus, comparisons with well-established classical ML techniques are not yet entirely fair, highlighting potential topics for further investigation \cite{QML_challenges_Cerezo,QML_challenges_Perdomo}. Among these, when considering input classical data, it is necessary to express them through quantum formalism in a process known as embedding. Strategies such as quantum feature maps \cite{Embedding_1} and data re‑uploading \cite{Embedding_2} have been adopted; these approaches directly influence both the computational resource requirements (e.g., number of qubits and circuit depth) and the the expressive power of the final model \cite{Embedding_effects}.

This work presents an in-depth analysis of how the available embedding techniques influence the results achievable in performing the classification task using a \gls*{vqa}\cite{VQA_Cerezo} of reference. The study addresses the embedding of input classical data through quantum formalism, and how it influcences the classification performance of the model \gls*{vqa}.

In this work, two main categories of embedding strategies are analyzed, Angle and Amplitude (described also in \cite{schuld2018supervised}). The Amplitude encoding strategy consist of embedding the input data vector as the probability amplitude of the state vector. On the other hand, for the Angle encoding strategy the data are passed as the angle parameters for the rotation gates. Further details related to these encoding strategies are described in \autoref{sec:Enc}. The Angle encoding mechanism has been subjected to several algorithmic approaches to understand the possibility of a better strategy for performing the embedding task. 

The study has been conducted on two established real datasets, i.e., Wine\cite{misc_wine_109} and Diabetes \cite{diabetes_dataset}, while keeping a constant number of qubits to implement the data classification model. This latter choice has posed a constraint on the definition of the subsequent model of \gls*{vqc}, thereby ensuring that the results are independent of it and focus solely on the efficacy of the embedding.
The analysis seeks to identify the best embedding strategy for a given input dataset. To this end, a set of multiple random input transformations is considered to understand if one best performs independently from the input data. The experiments have been executed on a local computer using a simulator provided within the Pennylane Library \cite{bergholm2022pennylane}, which can be considered representative of any kind of quantum computer's ideal behavior.\\

The results obtained in this work demonstrate that the embedding circuit has a significant impact on the classification performance of the model. Consequently, it can be regarded as a hyper-parameter and optimized through a benchmarking process specific to the dataset of interest. Indeed, when evaluating performance on both datasets, the differences in various metrics (accuracy, balanced accuracy, recall, precision and F1-score) between the best and worst models varies from 10\% to 30\% on average, considering the same kind of architecture (the employing of the re-uploading technique, which is also described in \autoref{sec:Reup}, and the number of layers of the circuits).\\

In the literature, a few other works have explored the topic of quantum embedding in classification problems and how it influences the models' performance. In particular, they generally focus on a limited set of analyses, which do not provide a comprehensive understanding of the effects on real-world case applications. Fauzi R. et al. \cite{fauzi2022analysis} have conducted tests on the effects of four different encoding techniques, namely Angle, Amplitude, Instantaneous Quantum Polynomial (IQP), and Complex Entangled, on the Iris dataset. In this case, the author suggest that the encoding strategy influences the results, but differently on our work, they do not consider all the possible combinations of the Angle encoding strategy, limiting the analysis to a single example of Angle encoding and not taking into account the impact of the choice of the rotational gates on classification performance.

On the other hand, Sierra-Sosa et al., in two different works\cite{sierra2020tensorflow,sierra2023data}, performed similar analyses but on a synthetic dataset. In the former case, the Authors compared the effects of applying Amplitude and Angle encoding strategies, while in the latter, Amplitude, IQP, and Second-order Pauli-Z evolution have been studied. In these cases, the Amplitude encoding strategies have obtained better results with respect the Angle encoding counterpart.
A. Matic et al. \cite{9951255} propose and analyze Quantum Convolutional Neural Networks (QCNNs) by varying both the encoding methods—focusing on high-order encoding and threshold encoding (embedding strategies that are also described in their article)—and the ansatz, including Basic and Strongly Entangling layers \cite{Schuld-basics-2}. Their study compares the performance of quantum models against classical counterparts, showing that the QCNN using high-order encoding and Basic Entangling layers achieved performance comparable to classical models.

Similarly, in \cite{monnet2024understandingeffectsdataencoding}, an exploration of Angle encoding models using RY and RX gates, Amplitude encoding models, and High-order models is presented. The study tested these models on two datasets, concluding that the encoding circuit influences the classification performance of quantum models. This work focuses exclusively on RX and RY Angle encoding models, excluding other combinations that are discussed and implemented in this article.

In all these works, the different encoding strategies have been applied without considering limitations on the number of qubits, which can significantly impact the resources required by the \gls*{vqa}. The outcome depends not only on the type of embedding technique used but also on the details of the related \gls*{qml} models, which vary in size and number of parameters. It is important to note that comparing results obtained from synthetic datasets with those already present in the state-of-the-art can be challenging. Additionally, when implementing the embedding, the default versions provided by two of the foremost open-source libraries (i.e., Qiskit and Pennylane) are adopted. This approach excludes the possibility of customizing the transformations applied to identify the most suitable ones for the classification task, especially in the case of the Angle encoding.\\

In summary, this paper proposes the following contributions:
\begin{itemize}
    \item Propose the evaluation of different Angle encoding strategies by tuning the rotational gates applied.
    \item Evaluate these Angle encoding strategies based on their trajectories on the Bloch sphere.
    \item Compare Angle encoding models with Amplitude encoding models using the same number of qubits (and thus the same number of parameters).
    \item Introduce a methodology for selecting the optimal VQC model.
\end{itemize}

In the following sections, the background topics of \gls*{ml} and \gls*{qc} will be presented, together with their intersection, \gls*{qml}. Then, the methodology behind the test conducted will be discussed in detail before providing the analysis results.

\section*{List of Abbreviations}
\begin{table}[ht]
  \centering
  \begin{tabular}{@{}ll@{}}
    \toprule
    \textbf{Abbreviazione} & \textbf{Significato} \\
    \midrule
    QC  & Quantum Computing \\
    QML & Quantum Machine Learning \\
    ML  & Machine Learning \\
    VQA & Variational Quantum Algorithm \\
    VQC & Variational Quantum Circuit \\
    \bottomrule
  \end{tabular}
  \caption{Table of abbreviations}
  \label{tab:abbreviation}
\end{table}

\section{Theoretical Foundation}
\gls*{qml} is an emerging research field based on developing models for \gls*{ml} by exploiting quantum technology to accelerate the training process and be capable of obtaining better results. The following sections will provide a detailed description of the two main topics, \gls*{qc} and \gls*{ml}, to ensure an accessible discussion to people with different backgrounds.
Further elements can be deepened respectively in \cite{nielsen2010quantum} and \cite{Goodfellow-et-al-2016}.

\subsection{Quantum Computing}
Quantum computing is a new computing paradigm that exploits quantum phenomena such as superposition and entanglement trying to reduce the processing time of specific algorithms.\\ 
The basic unit of quantum information is the qubit, which, unlike classical bits that can represent a value of either 0 or 1 at a time, can express a mixed state resulting from the linear combination of the two basis states $\ket{0}$ and $\ket{1}$. In particular, the generic state of the qubit $\ket{\psi}$ can be defined as:
\begin{equation}\label{eq:qubit}
    \ket{\psi} = \begin{pmatrix}
        c_0 \\ c_1
    \end{pmatrix} = c_0 \ket{0} + c_1 \ket{1}
\end{equation}
where $c_0$ is the probability amplitude associated with state $\ket{0}$, and $c_1$ the one associated with $\ket{1}$. As these coefficients are related to probability information, the norm of the resulting vector must be equal to 1:
\begin{equation}
  |c_0|^2 + |c_1|^2 = 1  
\end{equation}

By putting together a set of N qubits, a \gls*{qc} system is defined whose overall state can be expressed by a state vector $\ket{\psi}$ representative of a linear combination of $2^N$ basis states: 
\begin{equation} \label{eq:MultiqubitSystem}
    \ket{\psi} = \sum_{i=0}^{2^N - 1} c_i \ket{i}
\end{equation}
where $c_i$ is the probability amplitude associated with a particular state $\ket{i}$.\\

In order to perform computation, a series of transformations should be applied to the set of input qubits to modify the associated quantum state and to amplify the probability of the state corresponding to the solution of the problem under study. This series is called a quantum algorithm.\\
As it happens in the classical domain, where even complex algorithms are implemented through a sequence of simple basic operations expressed as electronic gates, in \gls*{qc} there is their counterpart, the quantum gates. The related transformations are mathematically described using a $2^n \times 2^n$ unitary matrix, where $n \leq N$ is the number of qubits to which the quantum gate is applied.
A notable example of single-qubit quantum transformations is represented by the Pauli gates, whose matrices associated are reported in \autoref{eq:Pauli}.
\begin{equation}
\begin{aligned}
X &= \begin{pmatrix} 0 & 1 \\ 1 & 0 \end{pmatrix} &
Y &= \begin{pmatrix} 0 & -i \\ i & 0 \end{pmatrix} &
Z &= \begin{pmatrix} 1 & 0 \\ 0 & -1 \end{pmatrix}
\end{aligned}\label{eq:Pauli}
\end{equation}

From the definition of these transformations, other well-known examples of single-qubit quantum gates can be derived, and, in particular, the rotational gates, which are mostly used for \gls*{qml} applications. Their effect on the reference qubit can be derived by considering the matrix exponential with the Pauli matrices as the exponents, as shown in \autoref{eq:RotationalGates}. They can also be defined by making explicit the angle parameter $\theta$, which describes the entity of the phase rotations applied to the qubit.
\begin{equation}
    \begin{aligned}
R_x(\theta) &= e^{-i\theta X / 2}  = \begin{pmatrix} \cos (\frac{\theta}{2}) & -i \sin (\frac{\theta}{2})  \\ -i \sin (\frac{\theta}{2}) & \cos (\frac{\theta}{2}) \end{pmatrix} \\
R_y(\theta) &= e^{-i\theta Y / 2}  = \begin{pmatrix} \cos (\frac{\theta}{2}) & -\sin (\frac{\theta}{2})  \\ \sin (\frac{\theta}{2}) & \cos (\frac{\theta}{2}) \end{pmatrix} \\ 
R_z(\theta) &= e^{-i\theta Z / 2}  = \begin{pmatrix} e^{-i \frac{\theta}{2}} & 0  \\ 0 & e^{i \frac{\theta}{2}} \end{pmatrix} 
\end{aligned}\label{eq:RotationalGates}
\end{equation}

Another well-known single-qubit gate is the Hadamard gate. It permits to obtain a quantum state in a uniform superposition between the two basis states $\ket{0}$ and $\ket{1}$. The matrix associated is:
\begin{equation} \label{eq:Hadamard}
    H = \frac{1}{\sqrt{2}} \begin{pmatrix} 1 & 1 \\ 1 & -1 \end{pmatrix}
\end{equation}

\begin{figure}
    \centering
    \includegraphics[width=0.3\linewidth]{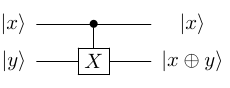}
    \caption{Graphical representation of the CNOT quantum gate, where $\ket{x}$ represents the control qubit while $\ket{y}$ the target one.}
    \label{fig:CNOT}
\end{figure}

Finally, the last quantum gate discussed in this brief introduction is the CNOT, shown in \autoref{fig:CNOT}, which, differently from the previously presented ones, acts on a couple of qubits (namely, the control and target). In particular, it implements a Pauli X transformation on the target qubit only when the control qubit is asserted. 
The CNOT gate creates entanglement between pairs of qubits, which is a quantum phenomenon where two or more particles become correlated in such a way that the quantum state of each particle cannot be described independently of the state of the other.

If two particles are entangled, knowing the state of one particle allows us to know the state of the other, no matter how far apart they are. This correlation occurs because the particles share a combined quantum state, not independent ones.

For two quantum systems $A$ and $B$, the general state of the combined system is described by a quantum state $\ket{\Psi}$ in the tensor product space of the two Hilbert space, denoted as $\mathcal{H}_A \otimes \mathcal{H}_B$.

If the system is not entangled, the total state can be written as the product of the states of each subsystem:
\begin{equation}
\ket{\Psi} = \ket{\psi_A} \otimes \ket{\psi_B}    
\end{equation}
where $\ket{\psi_A}$ is the state of system $A$ and $\ket{\psi_B}$ is the state of system $B$.
However, if the system is entangled, the state of the system cannot be factored into individual states of subsystems.

The matrix associated with the CNOT is:
\begin{equation}
    CNOT=
                    \begin{pmatrix}
                        1 & 0 & 0 & 0\\
                        0 & 1 & 0 & 0\\
                        0 & 0 & 0 & 1\\
                        0 & 0 & 1 & 0\\
                    \end{pmatrix}
\end{equation}

\subsection{Machine Learning}
\begin{figure*}
     \centering
     \begin{subfigure}[b]{0.32\textwidth}
         \centering
         \includegraphics[height=3.3cm]{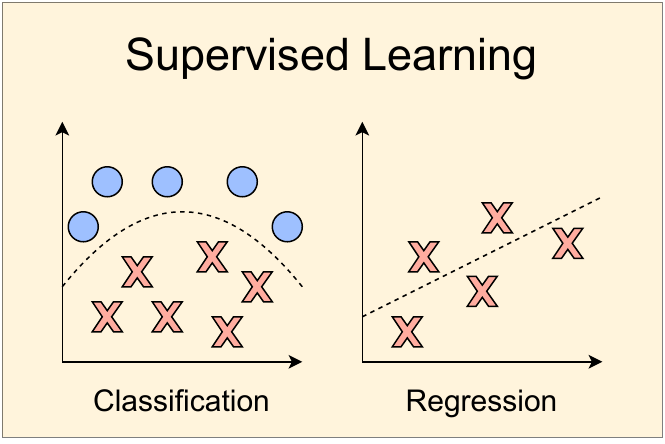}
         \caption{}
         \label{fig:ML-SL}
     \end{subfigure}
     \hfill
     \begin{subfigure}[b]{0.32\textwidth}
         \centering
         \includegraphics[height=3.3cm]{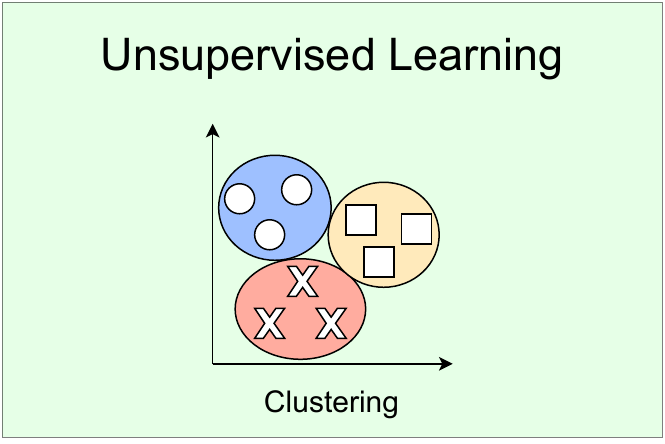}
         \caption{}
         \label{fig:ML-UL}
     \end{subfigure}
     \hfill
     \begin{subfigure}[b]{0.32\textwidth}
         \centering
         \includegraphics[height=3.3cm]{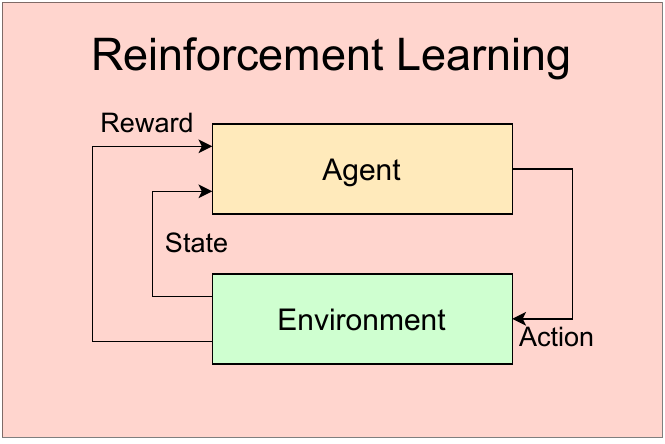}
         \caption{}
         \label{fig:ML-RL}
     \end{subfigure}
        \caption{Graphical representation of the typical subdivision of \gls*{ml} mechanisms, and application examples. \hyperref[fig:ML-SL]{(a)} Supervised learning algorithms, which are typically considered to perform \textit{data classification} or \textit{regression}. \hyperref[fig:ML-UL]{(b)} Unsupervised learning, where \textit{clustering} is the most common implementation. \hyperref[fig:ML-RL]{(c)} Reinforcement learning, whose main application is \textit{autonomous driving}.}
        \label{fig:TypeLearning}
\end{figure*}

\gls*{ml} is a branch of Artificial Intelligence that focuses on creating algorithmic models that can adapt their behavior based on the input data and a specific target. In a sense, these models learn from the input information during a training procedure to take actions coherently with expected outcomes. \gls*{ml} algorithms can be distinguished based on different learning mechanisms, with supervised, unsupervised, and reinforcement learning being the three primary categories.\\
Supervised learning algorithms are trained on structured input data, with explicit and labeled features provided by humans to identify these characteristics in new data. Unsupervised learning techniques consider unlabelled data to explore their characteristics and identify useful patterns for grouping purposes. Finally, reinforcement learning algorithms consider dynamic unlabelled data and aim to train agents to make efficient decisions autonomously in specific environments. Examples of possible applications for \gls*{ml} algorithms belonging to the three categories proposed are reported in \autoref{fig:TypeLearning}.\\

The current study aims to evaluate the influence of embedding techniques on implementing a supervised learning mechanism. Specifically, this analysis will address the data classification task, and the subsequent discussion will focus on its foundational principles.

\subsubsection{Classification Task}
In the domain of Supervised \gls*{ml} problems, a classification algorithm endeavors to establish a model that can assign input data to one or more predefined classes, identified by specific labels, based on a set of features. Broadly speaking, there are three distinct types of classification algorithms: \textit{binary}, \textit{multi-class}, and \textit{multi-label}, which vary in terms of the number and kind of output labels that can be assigned to the input data. The objective of binary classification is to distinguish data into one of the two mutually exclusive classes. On the other hand, multi-class and multi-label classifications are characterized by more than two possible classes to which the data can be assigned. However, in the former case, these classes are always mutually exclusive, whereas in the latter, this condition is less stringent.\\

\begin{figure}[h!]
    \centering
    \includegraphics[width=0.5\linewidth]{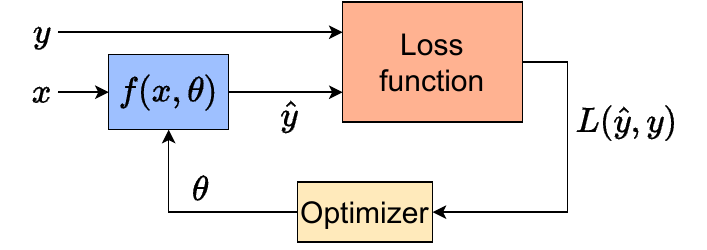}
    \caption{Graphical scheme of the training processing adopted for classification purposes. The model (i.e., the blue rectangle) receives the input data $x$ and estimates the output class they belong to, $\hat{y}$. The loss function then uses this information to evaluate the difference with respect to the correct target $y$, generating the magnitude of the error made by the model, $L(\hat{y},y)$. Finally, the optimizer closes the loop by considering the actual error and adjusting the parameters $\theta$ of the model to increase its prediction performance during the next iteration.}
    \label{fig:trainingModel}

\end{figure}

Focusing on the implementation details of such algorithms, it is always necessary to formulate a model composed of a parametric function $f(x,\theta)$ which can map input data $x$ to available output classes $y$. Once defined, the model can be trained using a loss function to evaluate the discrepancy between the actual output of the function $\hat{y}$, and its expected value, $y$. Additionally, an optimizer is employed to adjust the $\theta$ parameters of the classification model to reduce the gap between  $\hat{y}$ and $y$, effectively minimizing the classification error. A scheme of the overall process is presented in \autoref{fig:trainingModel}. \\

Several metrics can be utilized to assess the quality of a trained model on a specific dataset. One of the most important is \textit{accuracy}, which represents the percentage of correctly predicted samples over the entire dataset. For instance, in binary classification, where there are only two mutually exclusive classes (denoted as +1 and -1), the accuracy is calculated as:

\begin{equation} \label{eq:Acc}
    \mathrm{Accuracy} = \frac{\mathrm{TP} + \mathrm{TN}}{\mathrm{TP} + \mathrm{TN} + \mathrm{FP} + \mathrm{FN}} 
\end{equation}

\noindent
where:

\begin{conditions*}
    TP \ (True \ Positive)  &   Number of elements correctly predicted as positive\\
    TN \ (True \ Negative)  &   Number of elements correctly predicted as negative \\
    FP \ (False \ Positive) &   Number of elements miss predicted as positive\\
    FN \ (False \ Negative) &   Number of elements miss predicted as negative \\
\end{conditions*}

Other metrics, such as \textit{precision}, \textit{recall}, \textit{F1-score}, and \textit{Balanced accuracy} \cite{5597285} can be defined using the same notation. These are particularly useful for investigating the model performance on unbalanced datasets, i.e., where the distribution of the associated elements to the available classes is not uniform.
In particular, the precision represents the percentage of samples correctly predicted as true positive over all the data classified as positive:
\begin{equation} \label{eq:Precision}
    \mathrm{precision} = \frac{\mathrm{TP}}{\mathrm{TP} + \mathrm{FP}}
\end{equation}

\noindent
The recall is the percentage of data correctly predicted as positive over all the positive data:
\begin{equation} \label{eq:Recall}
    \mathrm{recall} = \frac{\mathrm{TP}}{\mathrm{TP} + \mathrm{FN}}
\end{equation}

\noindent
The F1-score is evaluated as the harmonic mean between precision and recall metrics, and it is defined as:
\begin{equation}\label{eq:F1}
    \mathrm{F1 \ score} = 2 * \frac{\mathrm{Precision} * \mathrm{Recall}}{\mathrm{Precision} + \mathrm{Recall}}
\end{equation}

The Balanced Accuracy is expressed as:
\begin{equation}
    \mathrm{Balanced \ accuracy} = \frac{\mathrm{Sensitivity + Specificity}}{2}
\end{equation}
where the \textit{Sensitivity} is the recall, while the \textit{Specificity} is the true negative rate, which is:
\begin{equation}
    \mathrm{Specificity} = \frac{TN}{TN + FP}
\end{equation}

\subsection{Quantum Machine Learning}
\label{sec:QML}
As introduced in the previous discussion, \gls*{qml} places itself in conjunction with two topics of great interest today: \gls*{qc} and \gls*{ml}. Among the approaches that can be found in the literature, the present work takes into consideration, for the related analyses, the particular model of \gls*{vqc} \cite{Farhi2018ClassificationWQ, PhysRevResearch.2.033125}.

\begin{figure}[h!]
    \centering
    \includegraphics[height=6cm]{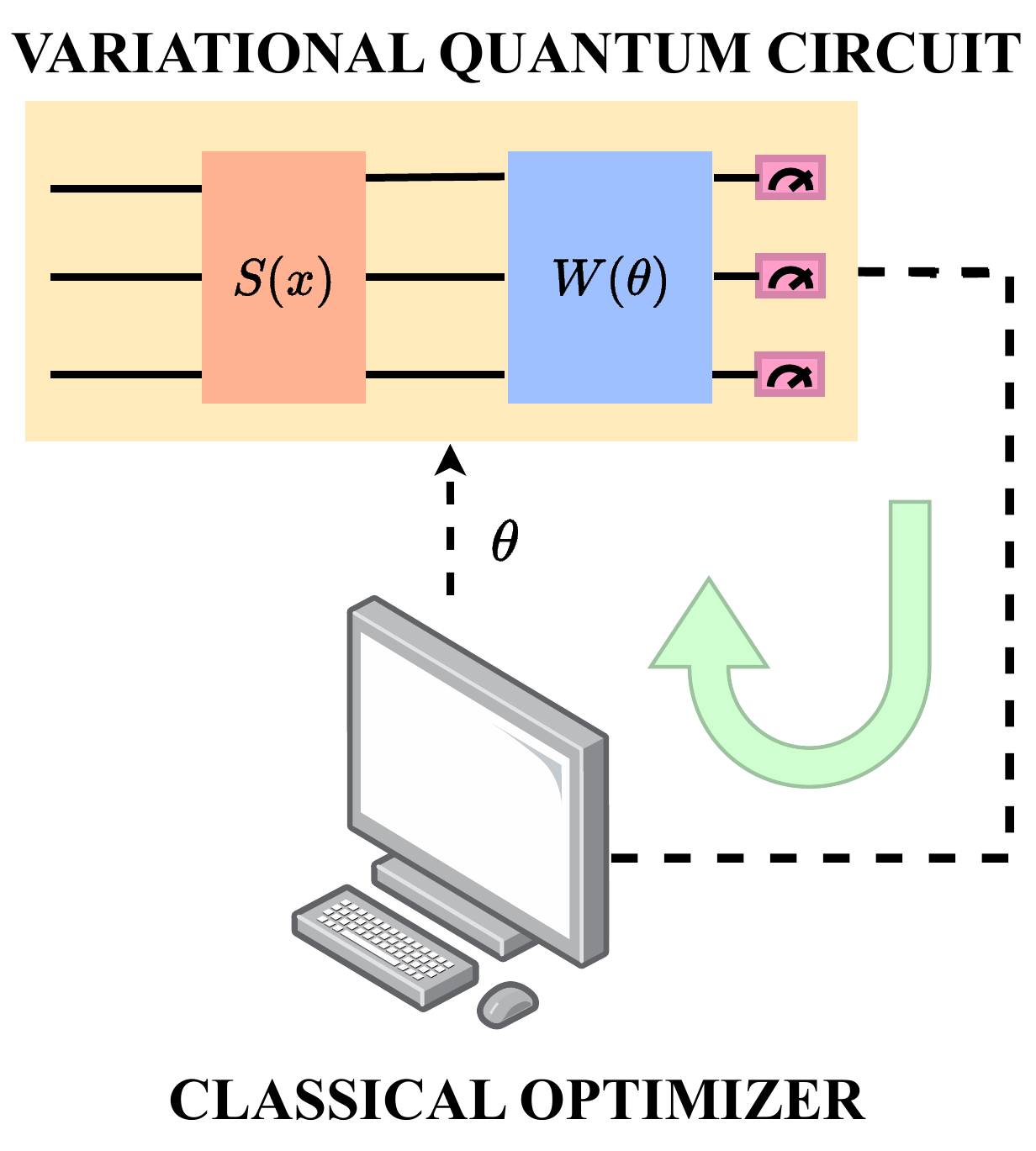}
    \caption{Representation of a training process for a \gls*{vqc}, composed of an encoding circuit, $S(x)$, and the ansatz, $W(\theta)$, used for the \gls*{ml} task. The output of the overall quantum circuit is then passed to a classical optimizer to tailor the angle parameters $\theta$ to reduce the classification error.}
    \label{fig:VariationalQuantumCircuit}
\end{figure}

This quantum algorithm, commonly referred to as the ansatz \cite{https://doi.org/10.1002/qute.201900070}, is composed of a series of parametric rotational gates whose values must be properly tuned during the training phase. Here, the evaluation of the best angles to be set is performed by a classical optimizer. This hybrid quantum-classical approach simplifies the implementation complexity on quantum machines, thus contributing to the model's success on the limited current quantum hardware. A general scheme of the \gls*{vqc} and its training procedure is shown in \autoref{fig:VariationalQuantumCircuit}, while a description of how the quantum circuit is constituted is discussed in the following sections.

\subsubsection{Encoding} \label{sec:Enc}
The encoding circuit is a fundamental circuit of the \gls*{vqc} that permits the representation of classical data following quantum formalism.
The input data provided to the \gls*{vqc} are represented by vectors whose elements are associated with the characteristics, called features, of the object considered.
Different approaches to implementing this circuit and some of them are addressed by the present work. In particular, the Basis, Amplitude, and Angle encoding. The details of these strategies are provided below, while for a wider perspective on the available embedding mechanisms, the reader is suggested to consider \cite{schuld2018supervised}, by Maria Schuld and Francesco Petruccione, in the chapter \textit{``Information Encoding"}.

\paragraph{Basis Encoding}
It is an embedding strategy that maps a features vector with N elements, each expressed with M bits, on a quantum circuit by placing  the different elements in superposition with each other in the state vector $\ket{\psi}$:
\begin{equation} \label{eq:SuperpositionBasisEnc}
  \ket{\psi} = \frac{1}{\sqrt{N}}\sum_{i=0}^{N-1} \ket{x_i}\ket{i}  
\end{equation}

\noindent
This strategy can be particularly effective if the number of features to be represented in the quantum circuit is larger than one because the number of qubit scales as in \autoref{eq:ScalingBasisEncoding}:
\begin{equation} \label{eq:ScalingBasisEncoding}
    \#\mathrm{qubit} = M + \log_2(N)
\end{equation}

\begin{figure*}[h!]
    \centering
    \includegraphics[height=4cm]{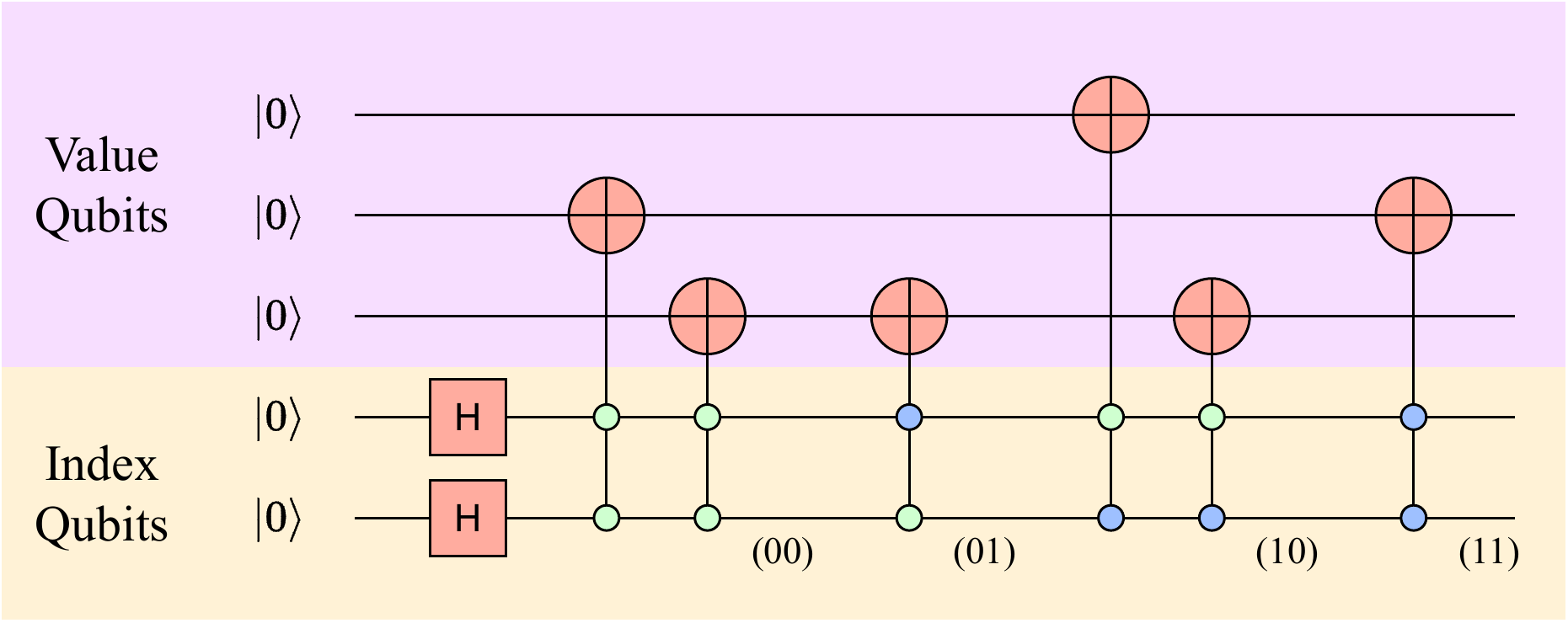}
    \caption{Representation of the Basis Encoding strategy in which the feature vector [011, 001, 101, 010] is embedded into the quantum circuit. Each feature is encoded using CNOT gates, whose control qubits are referenced to the feature index (blue circles when the control is active with input $\ket{1}$, green when it is $\ket{0}$).}
    \label{fig:BasisEncoding}
\end{figure*}

\noindent
\autoref{fig:BasisEncoding} shows an example of the application of the Basis Encoding to a feature vector to embed the elements [$011_{00}$, $001_{01}$, $101_{10}$, $010_{11}$].

\paragraph{Amplitude encoding}
In this case, the value of each element of the input data vector is transformed into an amplitude probability of the state vector associated with the quantum circuit. At first, to implement this technique, it is necessary to normalize the input vector $x$ by dividing each feature by the norm-2 of the same vector. Then, it is possible to map the vector as a state vector for the quantum circuit,
\begin{equation} \label{eq:SuperpositionAmplitudeEnc}
    \ket{\psi} = \frac{1}{\sqrt{N}} \sum_{i=0}^{N-1} x_i \ket{i}
\end{equation}

\begin{figure*}[h]
    \centering
    \includegraphics[height=4cm]{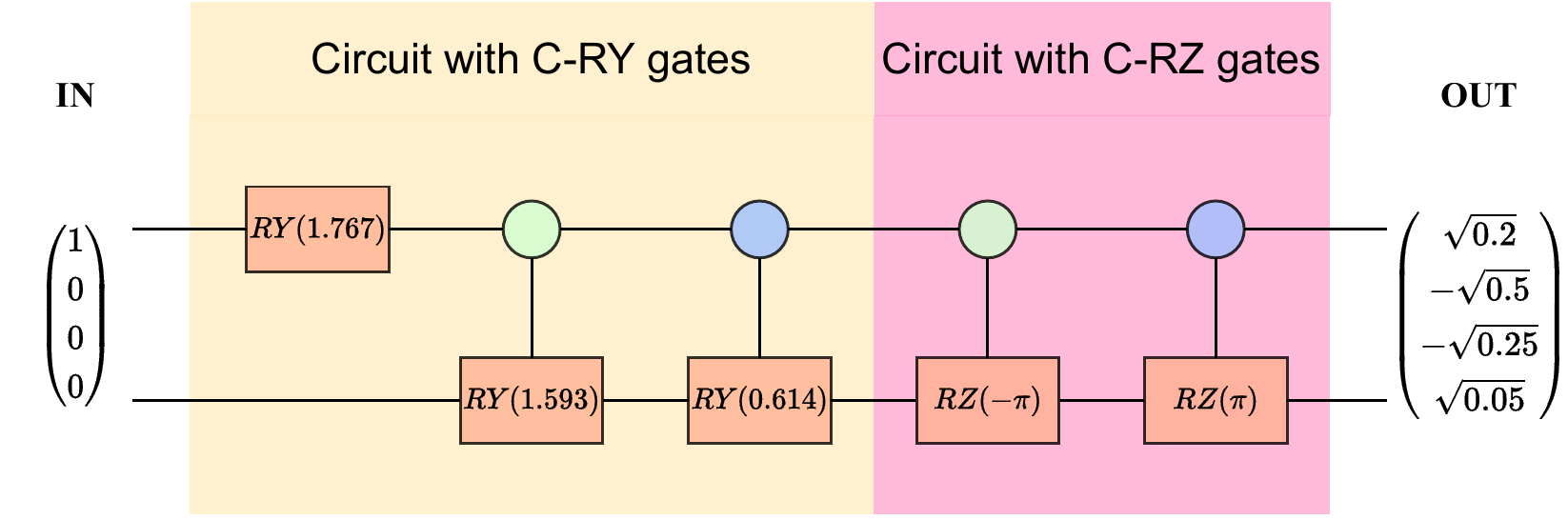}
    \caption{Representation of a Mottonen State Preparation circuit that encodes the state vector [$\sqrt{0.2}$, $-\sqrt{0.5}$, $-\sqrt{0.25}$, $\sqrt{0.05}$]. The controlled gates, whose circles are colored in green, are active when the control qubit is in state $\ket{0}$; otherwise, they are active when the control is in state $\ket{1}$.}
    \label{fig:MottonenComplete}
\end{figure*}

\noindent
In particular, the Amplitude encoding strategy can be implemented by exploiting the Mottonen State Preparation\cite{mottonen2004transformation}, which is composed of a sequence of two main blocks: a circuit made of controlled RY gates to modify the module of the amplitude probabilities of the state vector and a circuit composed of controlled RZ gates to change the phases of the belonging qubits. \autoref{fig:MottonenComplete} shows an example of an application.\\

Considering the Amplitude encoding requires a number of qubits which scales with the logarithm to the base 2 of the number of elements belonging to the data vector, 
\begin{equation} \label{eq:ScalingAmplitudeEncoding}
    \#\mathrm{qubit} = \log_2(N)
\end{equation}

\begin{figure}
    \centering
    \includegraphics[height=4cm]{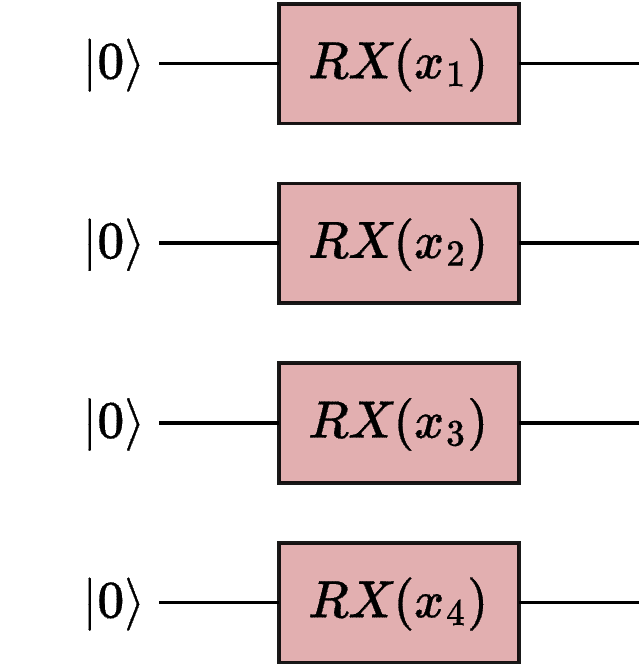}
    \caption{Representation of an Angle encoding strategy using RX as the rotational gate. Each feature of the input vector is embedded in a qubit of the quantum circuit.}
    \label{fig:AngleEncodingCircuitRX}
\end{figure}

\paragraph{Angle encoding}
It consists of encoding data as angles for rotational gates. In this case, each feature is encoded on a different qubit, and the final state following these transformations will be
\begin{equation} \label{eq:SuperpositionAngleEnc}
    \ket{\psi} = (\otimes_{i=0}^{N-1} R(x_i)) \ket{00...0} 
\end{equation}
An example of using the RX gate to embed the data into the quantum circuit is shown in  \autoref{fig:AngleEncodingCircuitRX}.\\

By employing the Angle encoding strategy, the number of the qubits needed scales as the number of features of the input data vector, \\
\begin{equation} \label{eq:ScalingAngleEncoding}
    \text{qubit} = N
\end{equation}

It is important to highlight that the most commonly employed strategy in the state-of-the-art involves using the RY gate for encoding data~\cite{Hur2022, GONG2024129993, https://doi.org/10.1049/qtc2.12032}.

\subsubsection{Ansatz}
The ansatz is the core of the \gls*{vqc}, which allows data processing to implement the classification task. It generally comprises two main components: a rotational circuit, constituted by rotational gates with parametric angles, and an entanglement circuit implemented with controlled gates. Since the choice of ansatz circuit topology directly influences the model’s classification performance, it represents as a key hyperparameter in defining the \gls*{qml} model.

\subsubsection{Re-uploading technique} \label{sec:Reup}
\begin{figure}[!h]
    \centering
    \includegraphics[width=0.5\linewidth]{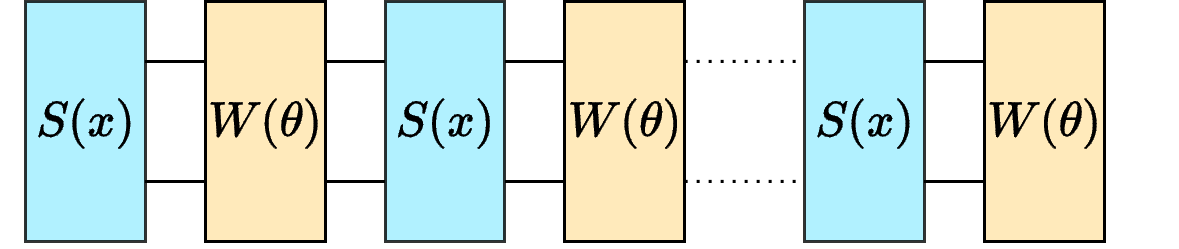}
    \caption{Example of exploiting the re-uploading technique. The blue rectangles represent the embedding circuits, while the yellow ones are the ansatze.}
    \label{fig:re-uploading}
\end{figure}

The re-uploading technique \cite{perez2020data}, illustrated in \autoref{fig:re-uploading}, is intended to improve the \gls*{qml} model's accuracy by embedding the input data multiple times in the quantum circuit, increasing thereby the model's expressive strength and ability to describe complicated data patterns.

\subsubsection{Measurement}
Once the quantum circuit has been defined, the measurement operation is performed to determine the class to which the data belongs. In particular, two different strategies are defined depending on whether the classification is binary or multi-class. For the former, the probability of a single qubit being in state $\ket{1}$ is measured. If it exceeds a threshold, typically $50\%$, the input data is assigned to class 1; otherwise, it is labeled as 0. For the latter, where each input data belongs to one of possible M classes, the probability of being in the $\ket{1}$ state is measured on the associated M qubits. In this case, after the measurement operation, classical post-processing is necessary, and the Softmax function is applied to the resulting probabilities $z_i$ of each i-th qubit:
\begin{equation}
\label{eq:Softmax}
    \sigma(z)_i = \frac{e^{z_i}}{\sum_{j=1}^M e^{z_j}}
\end{equation}

This function acts as a normalization operation, ensuring that the sum of the components of the output state vector equals 1. The input data is then assigned to the class associated with the highest probability qubit.

\section{Implementation}
\begin{figure*}[]
    \centering
    \includegraphics[height=4.5cm]{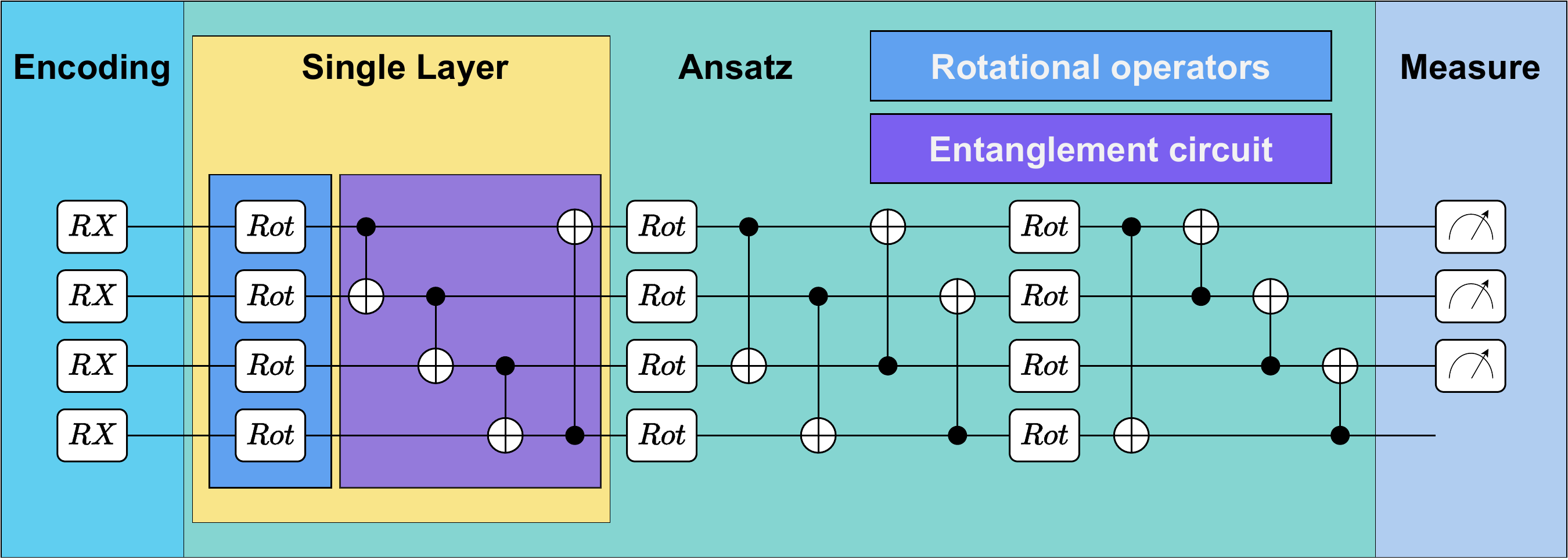}
    \caption{Representation of an example of a \gls*{vqc} model. The encoding circuit is realized using an Angle encoding technique with RX gates, while the Ansatz is the Strongly Entangling layer circuit \cite{Schuld-basics-2}, realized with rotational (ROT) and CNOT gates.}
    \label{fig:VQCLoop}
\end{figure*}

The proposed work focuses on the \gls*{vqc} and its main parameters, described in details in the \gls*{qml} \autoref{sec:QML}. Specifically, an analysis is conducted on the \textit{encoding mechanism}, the \textit{number of layers of the variational part}, and the application of the \textit{re-uploading strategy}. The study measures the accuracy achieved by modifying these parameters to identify the optimal \gls*{qml} model for the classification task under consideration.\\
An example of an implemented model is shown in \autoref{fig:VQCLoop}, where the most significant blocks that compound the quantum circuit, such as the encoding and the variational circuit, are highlighted.\\

This section discusses the actual implementation and the methodology used to test the models' performance, providing the necessary details for the reader to replicate the experiments.

\subsection{Encoding circuit}
As previously discussed in the background section, the encoding is a fundamental component of a quantum circuit that enables the representation of data in the quantum domain for processing. In the experiments conducted in this work, Angle and Amplitude encoding strategies have been implemented. These strategies are suitable for near-term application solutions as they can be applied to a limited number of qubits.\\

Before encoding data into the quantum circuit, a pre-processing step is necessary to normalize the input data vector in the range of variation of the quantities involved by the specific embedding mechanism, i.e., the phase or the amplitude probability of the qubits. The data is normalized in the interval [0, $\pi$] for both the Angle and Amplitude encoding strategies. An additional normalization is applied only to the latter by dividing the state vector by its norm-two.\\

\noindent
Regarding the implementation of the Amplitude encoding strategy, the Mottonen State Preparation, discussed in \autoref{sec:Enc}, has been adopted. \\

By contrast, Angle encoding offers a high degree of implementation flexibility: one may employ various combinations of rotational gates with the option of prepending a layer of Hadamard gates to prepare the input qubits in a uniform superposition.
Therefore, a set of possible combinations of embedding circuits for the Angle encoding strategy can be identified, and the methodology for choosing them is discussed in the following.\\
\subsubsection{Angle encoding circuits}
An initial set of feasible permutations has been defined, consisting of groups of sizes 1, 2, 3, and 4 of rotational quantum gates \textit{(i.e., $R_X$, $R_Y$, and $R_Z$)} preceded, if possible, by Hadamard transformations. Then, the equivalent circuits have been identified, and only one instance between them has been considered within the set. For example, the circuit's behavior when applying RZ-RX rotational gates or the RX gate alone is identical. In fact, applying the encoding circuits on a single qubit initially in the $\ket{0}$ state, the resulting state vectors are reported in \autoref{eq:RXEnc} and \autoref{eq:RZ-RXEnc} for respectively the RX and RZ-RX strategies.
\begin{equation} \label{eq:RXEnc}
    RX(x) \ket{0} = \begin{pmatrix}
        \cos(\frac{x}{2}) \\ -i\sin(\frac{x}{2})
    \end{pmatrix}
\end{equation}
\begin{equation} \label{eq:RZ-RXEnc}
    RX(x) RZ(x)\ket{0} = e^{-i\frac{x}{2}}\begin{pmatrix}
        \cos(\frac{x}{2}) \\ -i\sin(\frac{x}{2})
    \end{pmatrix}
\end{equation}

\noindent
Twenty possible Angle encoding strategies have been selected, as shown in \autoref{tab:EncodingStrat}. Unlike the majority of works presented in the state-of-the-art, as reported in the Introduction, where the RY strategy is predominantly employed, this paper explores the entire solution space of Angle encoding techniques to observe how they influence the quality of the model.

\begin{table}[h!]
  \centering
  \small
  \begin{tabular}{cccc}
    \toprule
    1-gate & 2-gates & 3-gates & 4-gates \\
    \midrule
    RX & RX-RY & RX-RY-RZ & H-RY-RX-RZ \\  
    RY & RX-RZ & RX-RZ-RY & H-RY-RZ-RX \\ 
    & RY-RX & RY-RX-RZ & H-RZ-RX-RY \\ 
    & RY-RZ & RY-RZ-RX & H-RZ-RY-RX \\ 
    & H-RY & H-RY-RX & \\ 
    & H-RZ & H-RY-RZ & \\ 
    & & H-RZ-RX & \\ 
    & & H-RZ-RY & \\
    \bottomrule
  \end{tabular}
  \caption{Angle encoding strategies grouped per number of transformations required.}
  \label{tab:EncodingStrat}
\end{table}

\begin{figure*}[h]
    \centering
    \begin{subfigure}[b]{0.3\textwidth}
        \centering
        \includegraphics[width=\textwidth]{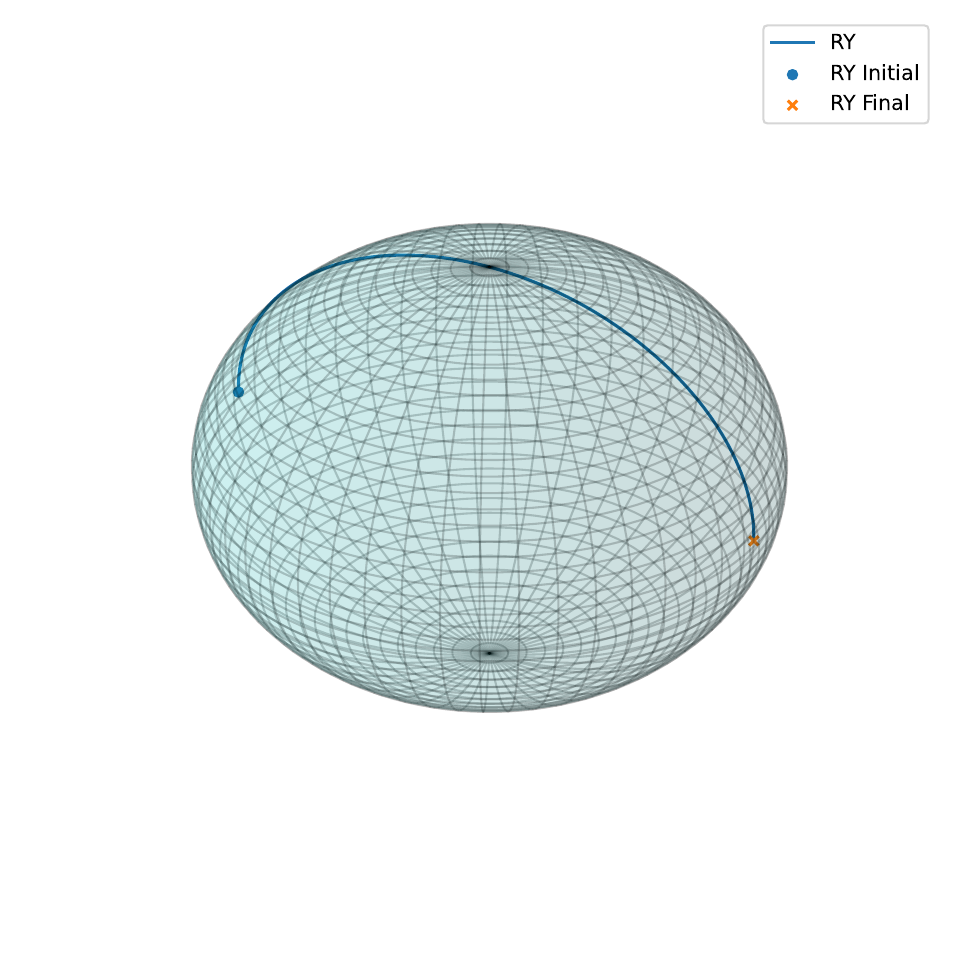}
        \caption{H-RX}
        \label{subfig:bloch_rx}
    \end{subfigure}
    \begin{subfigure}[b]{0.3\textwidth}
        \centering
        \includegraphics[width=\textwidth]{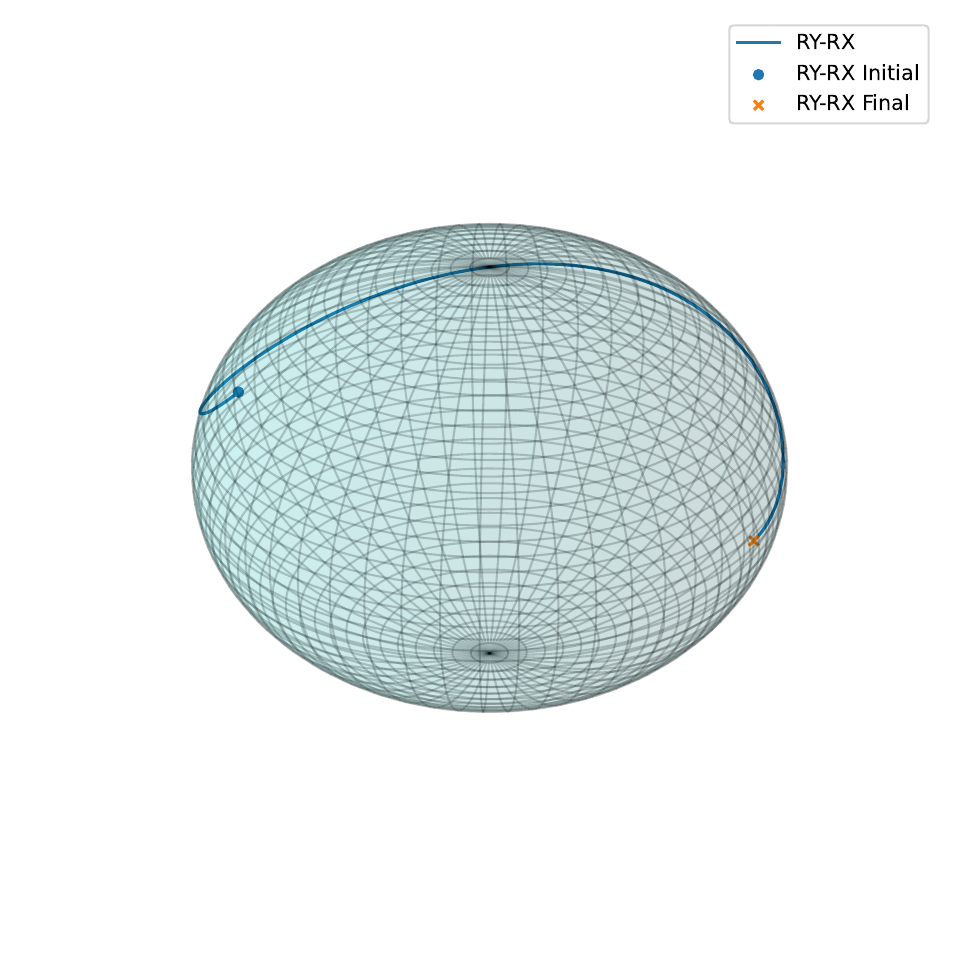}
        \caption{H-RY-RX}
        \label{subfig:bloch_rx_ry}
    \end{subfigure}
    \begin{subfigure}[b]{0.3\textwidth}
        \centering
        \includegraphics[width=\textwidth]{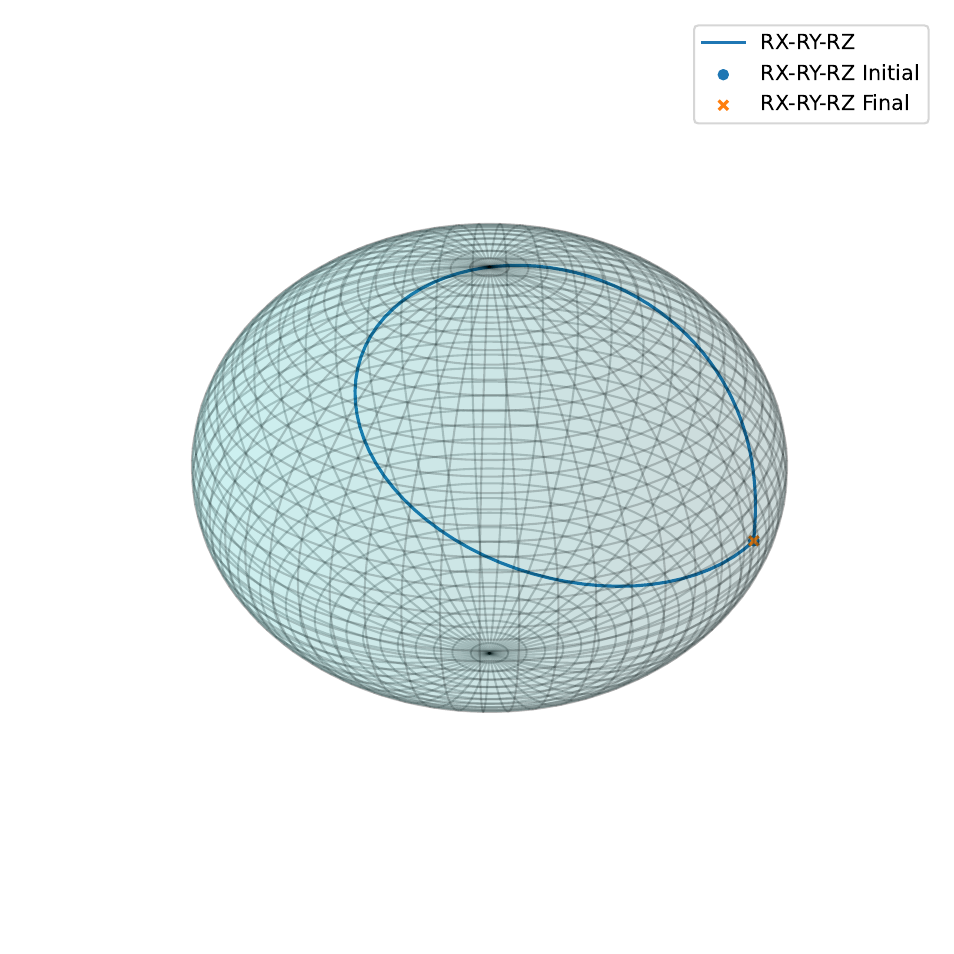}
        \caption{H-RY-RX-RZ}
        \label{subfig:bloch_rx_ry_rz}
    \end{subfigure}
    \begin{subfigure}[b]{0.3\textwidth}
        \centering
        \includegraphics[width=\textwidth]{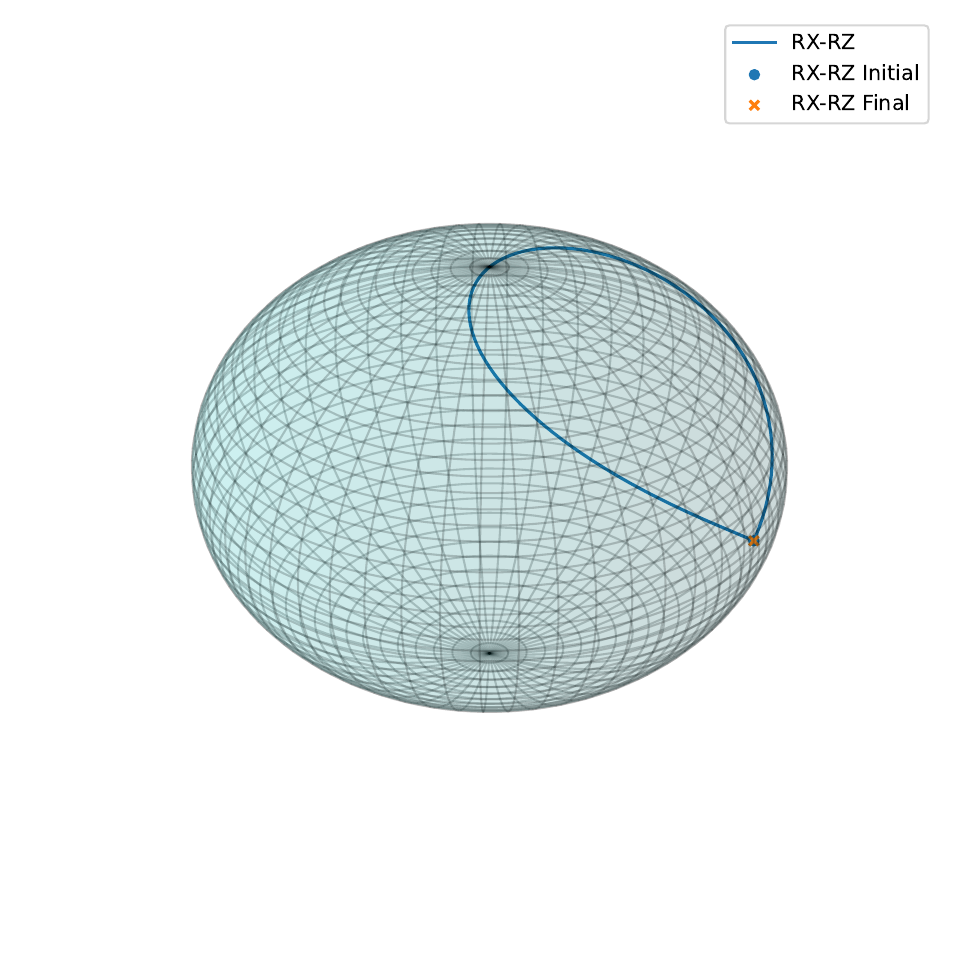}
        \caption{H-RY-RZ}
        \label{subfig:bloch_rx_rz}
    \end{subfigure}
    \begin{subfigure}[b]{0.3\textwidth}
        \centering
        \includegraphics[width=\textwidth]{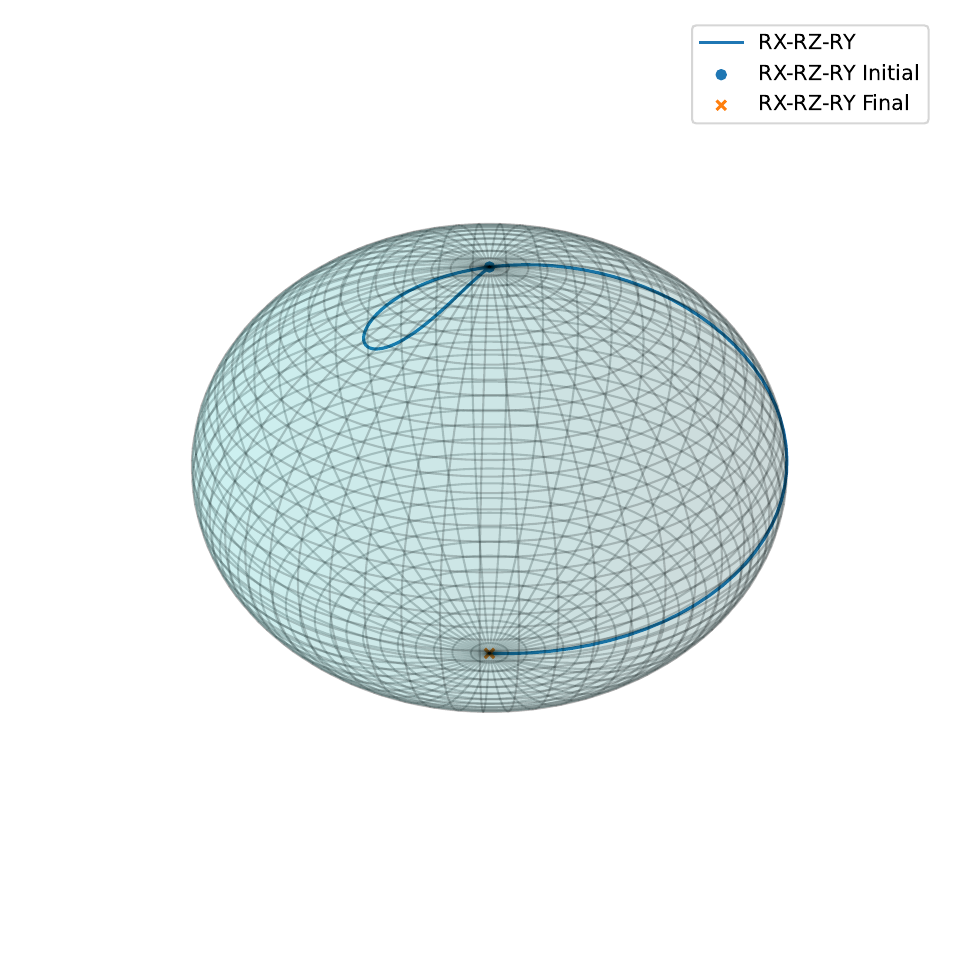}
        \caption{H-RY-RZ-RX}
        \label{subfig:bloch_rx_rz_ry}
    \end{subfigure}

\caption{Representations of the trajectories of applying the transformations RX, RX-RY, RX-RY-RZ, RX-RZ, and RX-RZ-RY varying the angle parameter in the interval [$-\pi / 2$, $\pi / 2$].}
\label{fig:bloch_start_rx}
\end{figure*}

\begin{figure*}[h]
    \centering
    \begin{subfigure}[b]{0.3\textwidth}
        \centering
        \includegraphics[width=\textwidth]{Images/BlochSphere/bloch_ry.pdf}
        \caption{H-RY}
        \label{subfig:bloch_ry}
    \end{subfigure}
    \begin{subfigure}[b]{0.3\textwidth}
        \centering
        \includegraphics[width=\textwidth]{Images/BlochSphere/bloch_ry_rx.pdf}
        \caption{H-RY-RX}
        \label{subfig:bloch_ry_rx}
    \end{subfigure}
    \begin{subfigure}[b]{0.3\textwidth}
        \centering
        \includegraphics[width=\textwidth]{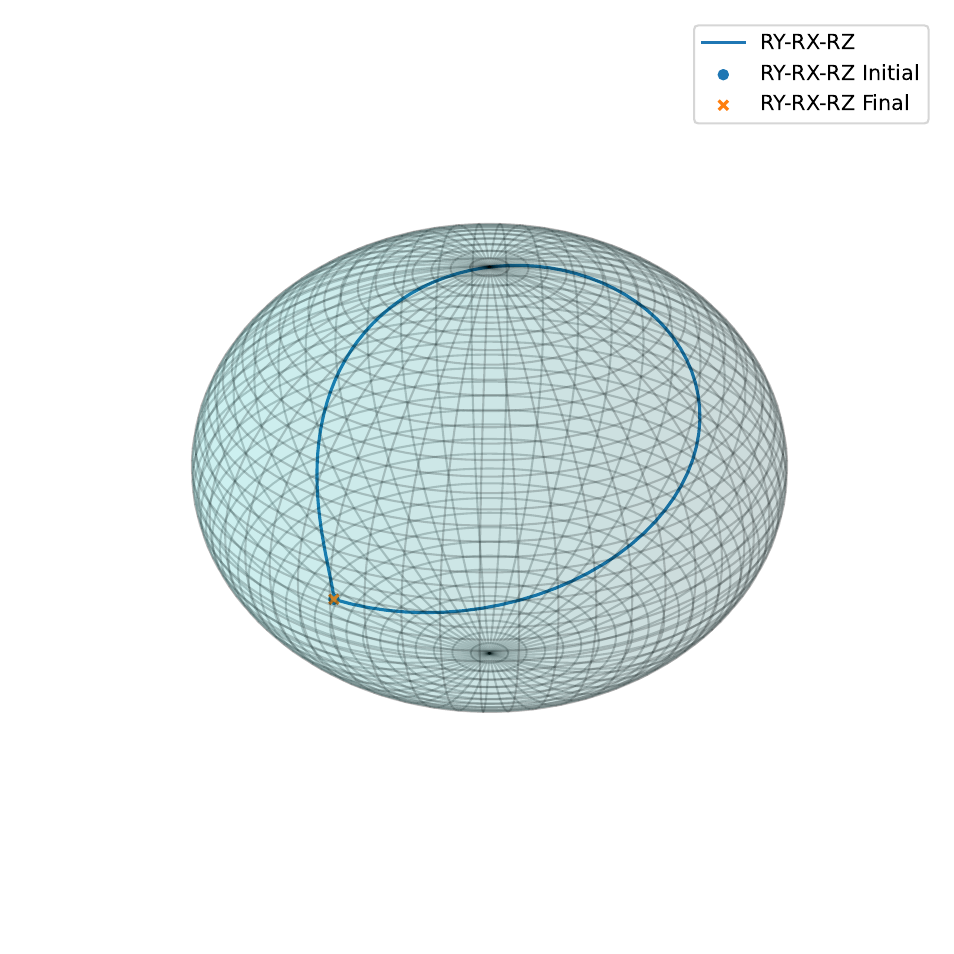}
        \caption{H-RY-RX-RZ}
        \label{subfig:bloch_ry_rx_rz}
    \end{subfigure}
    \begin{subfigure}[b]{0.3\textwidth}
        \centering
        \includegraphics[width=\textwidth]{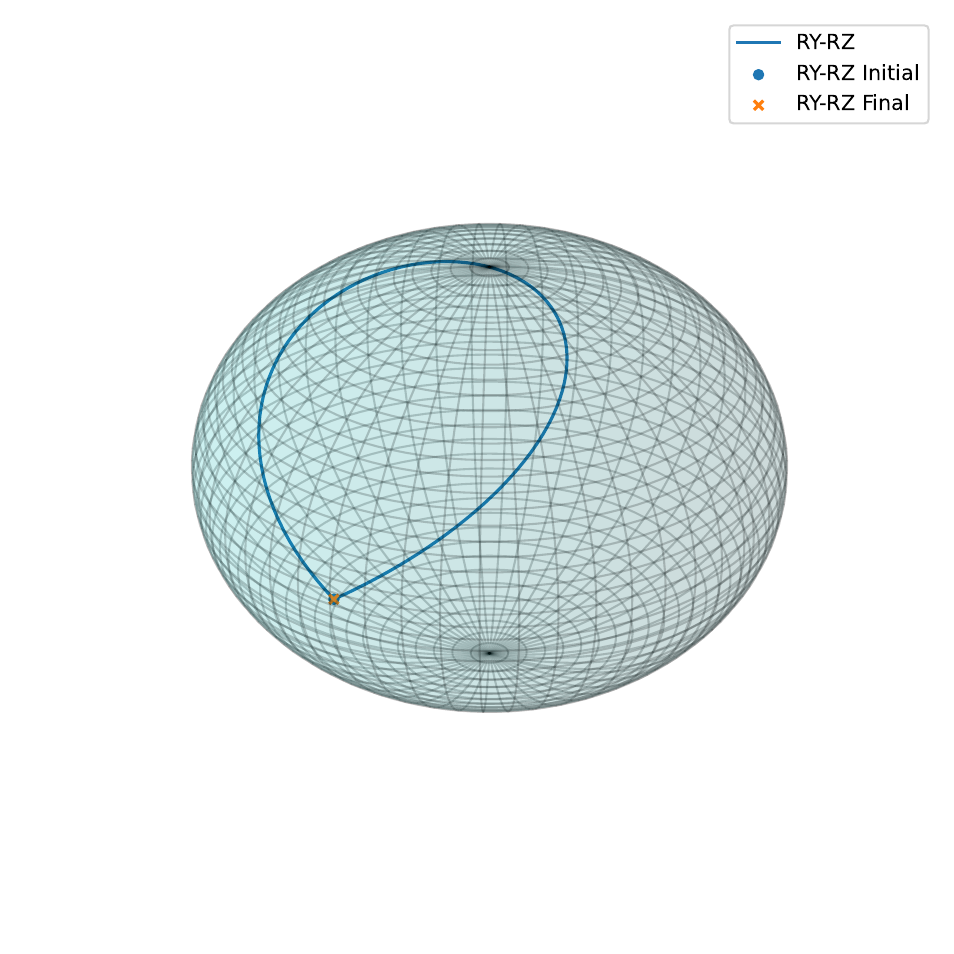}
        \caption{H-RY-RZ}
        \label{subfig:bloch_ry_rz}
    \end{subfigure}
    \begin{subfigure}[b]{0.3\textwidth}
        \centering
        \includegraphics[width=\textwidth]{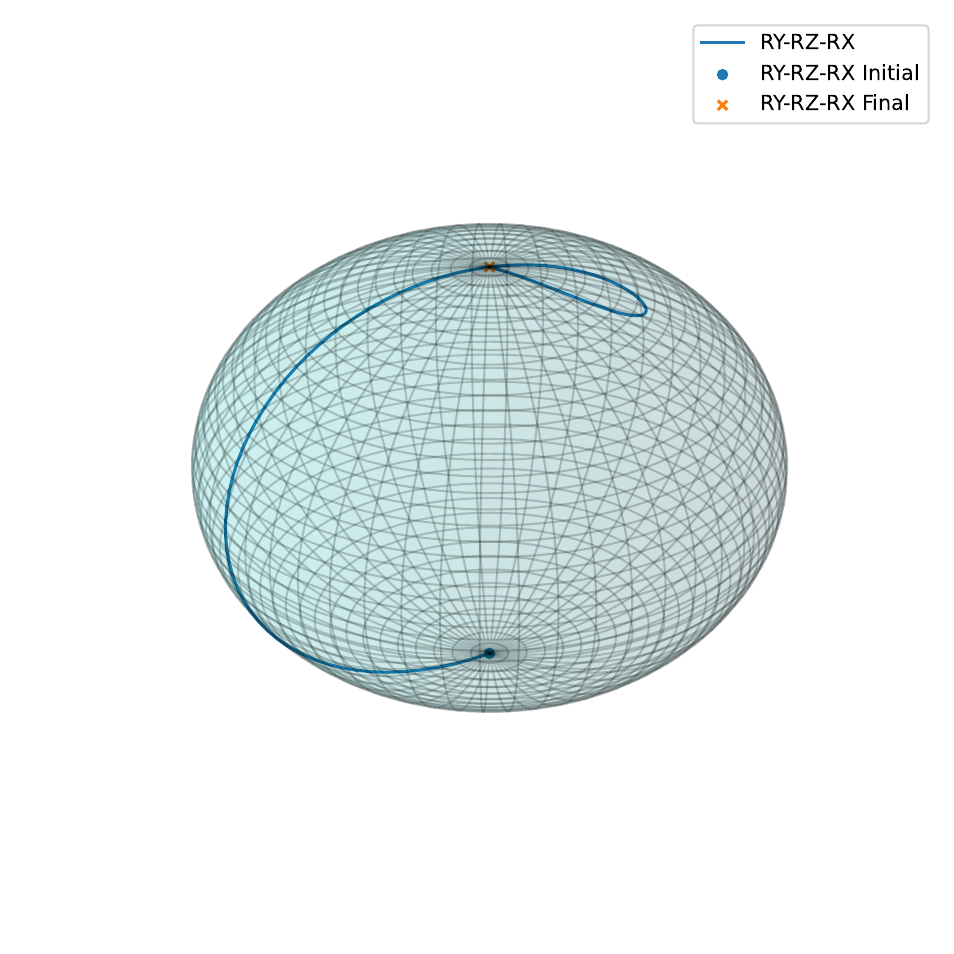}
        \caption{H-RY-RZ-RX}
        \label{subfig:bloch_ry_rz_rx}
    \end{subfigure}

\caption{Representations of the trajectories of applying the transformations H-RY, H-RY-RX, H-RY-RX-RZ, H-RY-RZ, and H-RY-RZ-RX varying the angle parameter in the interval [$-\pi / 2$, $\pi / 2$].}
\label{fig:bloch_start_ry}
\end{figure*}

\begin{figure*}[h]
    \centering
    \begin{subfigure}[b]{0.3\textwidth}
        \centering
        \includegraphics[width=\textwidth]{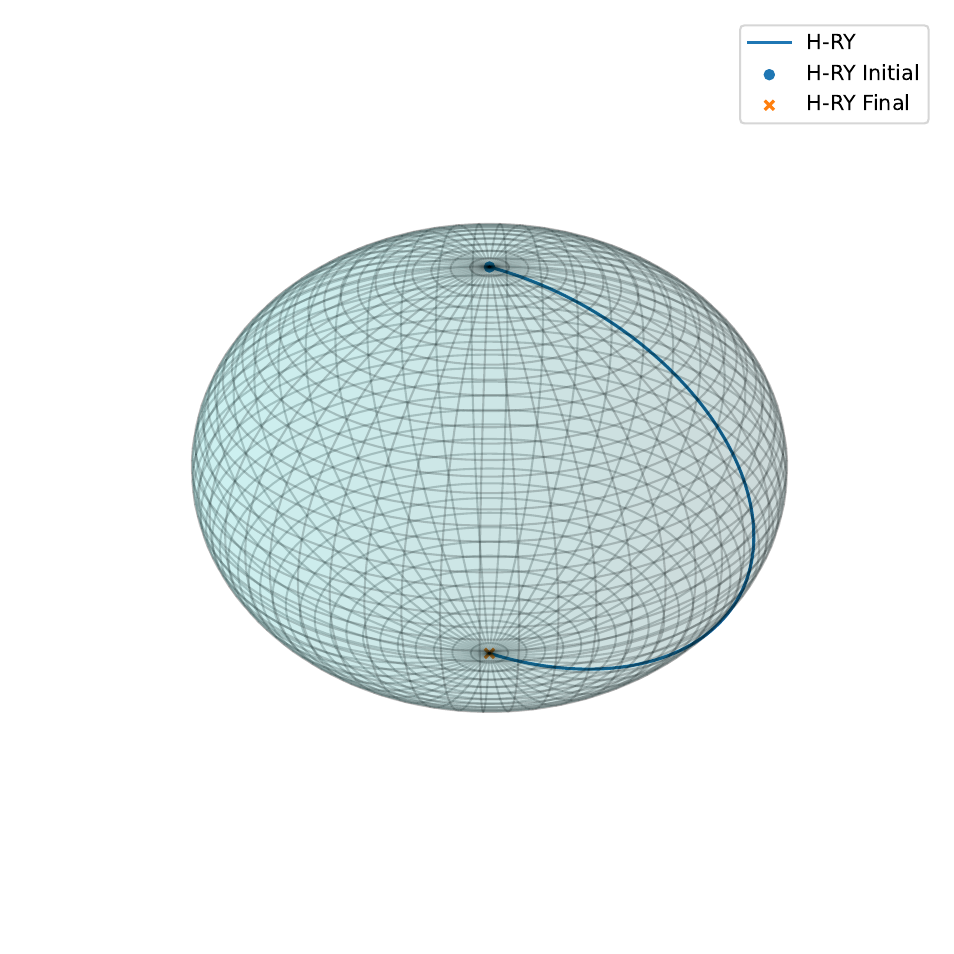}
        \caption{H-RY}
        \label{subfig:bloch_h_ry}
    \end{subfigure}
    \begin{subfigure}[b]{0.3\textwidth}
        \centering
        \includegraphics[width=\textwidth]{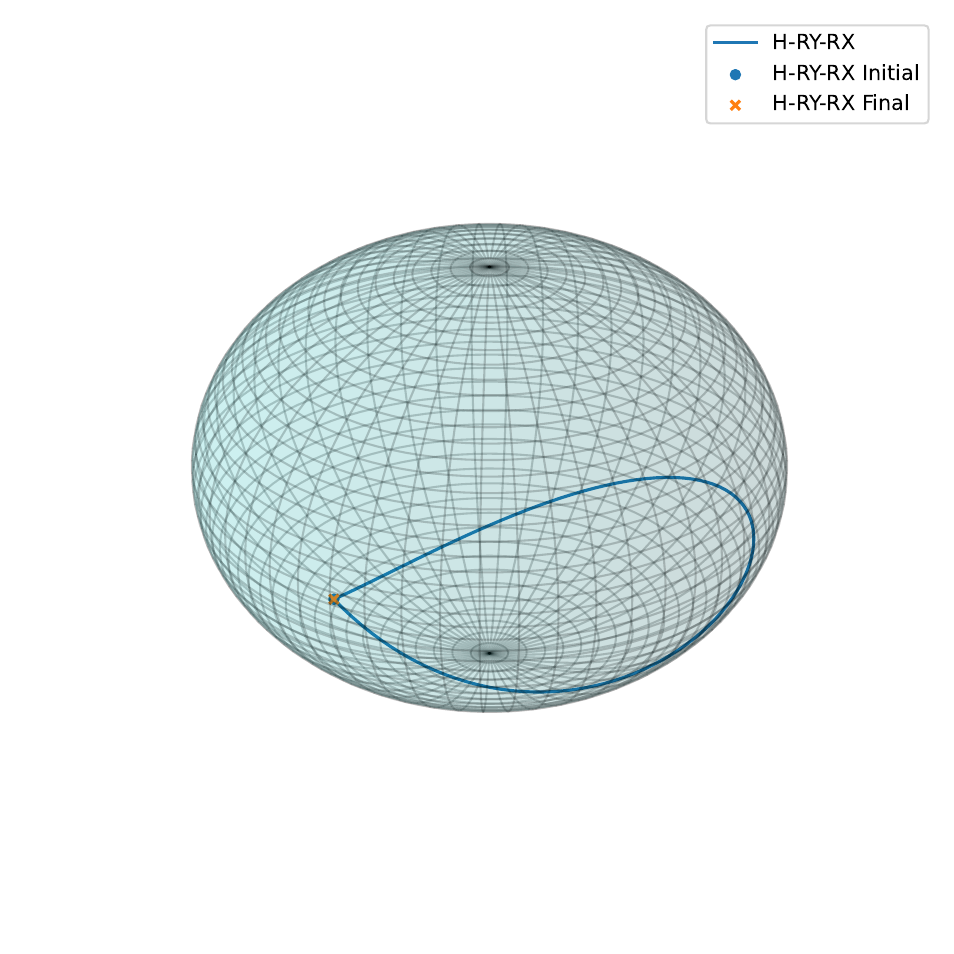}
        \caption{H-RY-RX}
        \label{subfig:bloch_h_ry_rx}
    \end{subfigure}
    \begin{subfigure}[b]{0.3\textwidth}
        \centering
        \includegraphics[width=\textwidth]{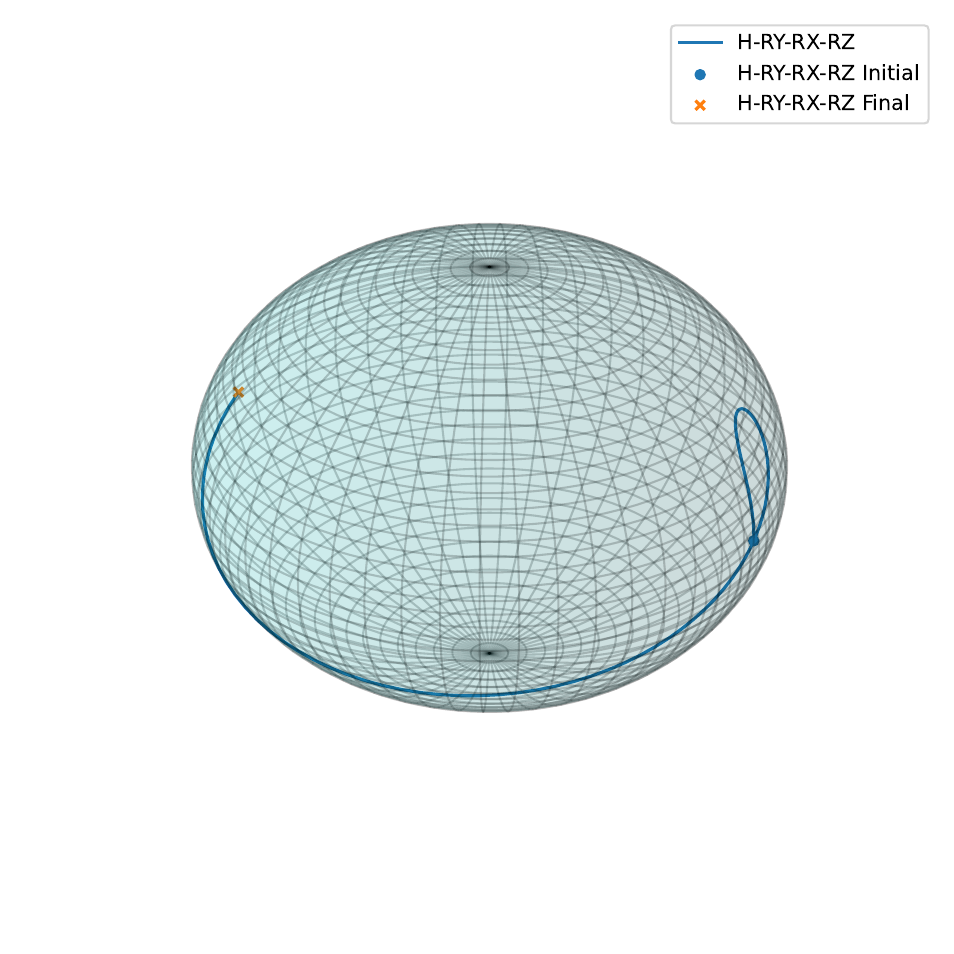}
        \caption{H-RY-RX-RZ}
        \label{subfig:bloch_h_ry_rx_rz}
    \end{subfigure}
    \begin{subfigure}[b]{0.3\textwidth}
        \centering
        \includegraphics[width=\textwidth]{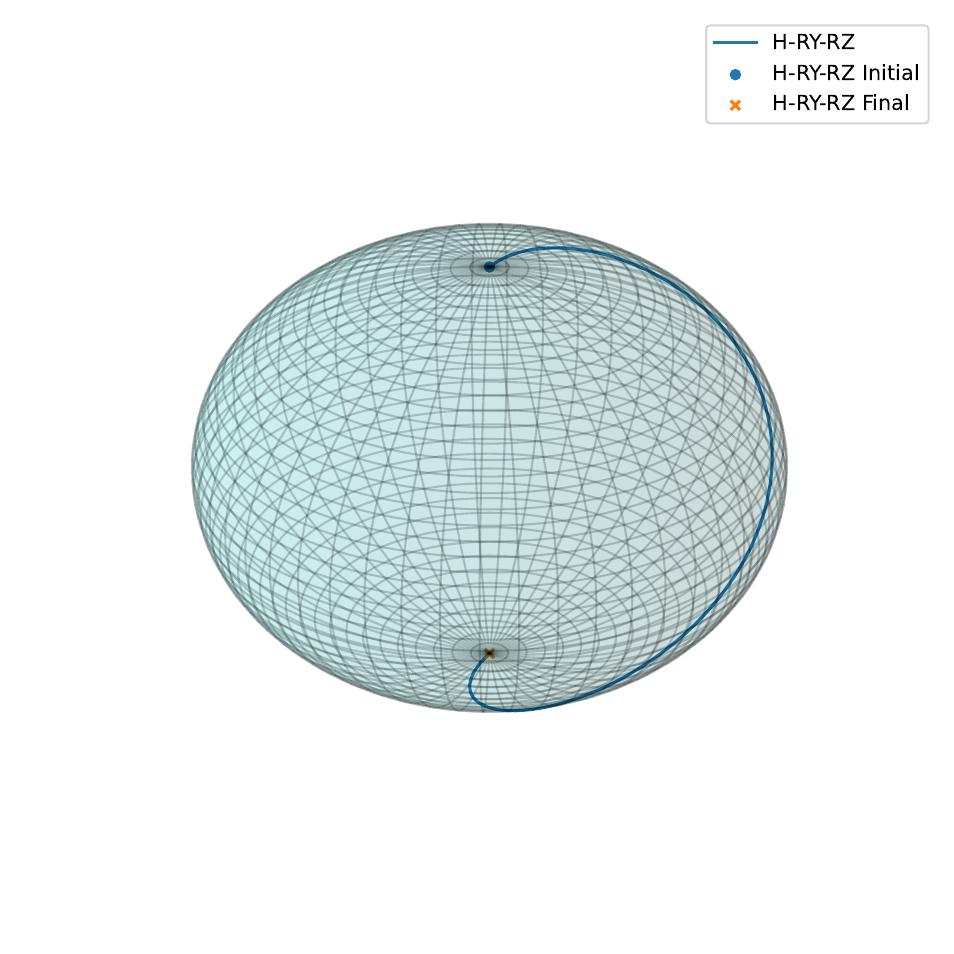}
        \caption{H-RY-RZ}
        \label{subfig:bloch_h_ry_rz}
    \end{subfigure}
    \begin{subfigure}[b]{0.3\textwidth}
        \centering
        \includegraphics[width=\textwidth]{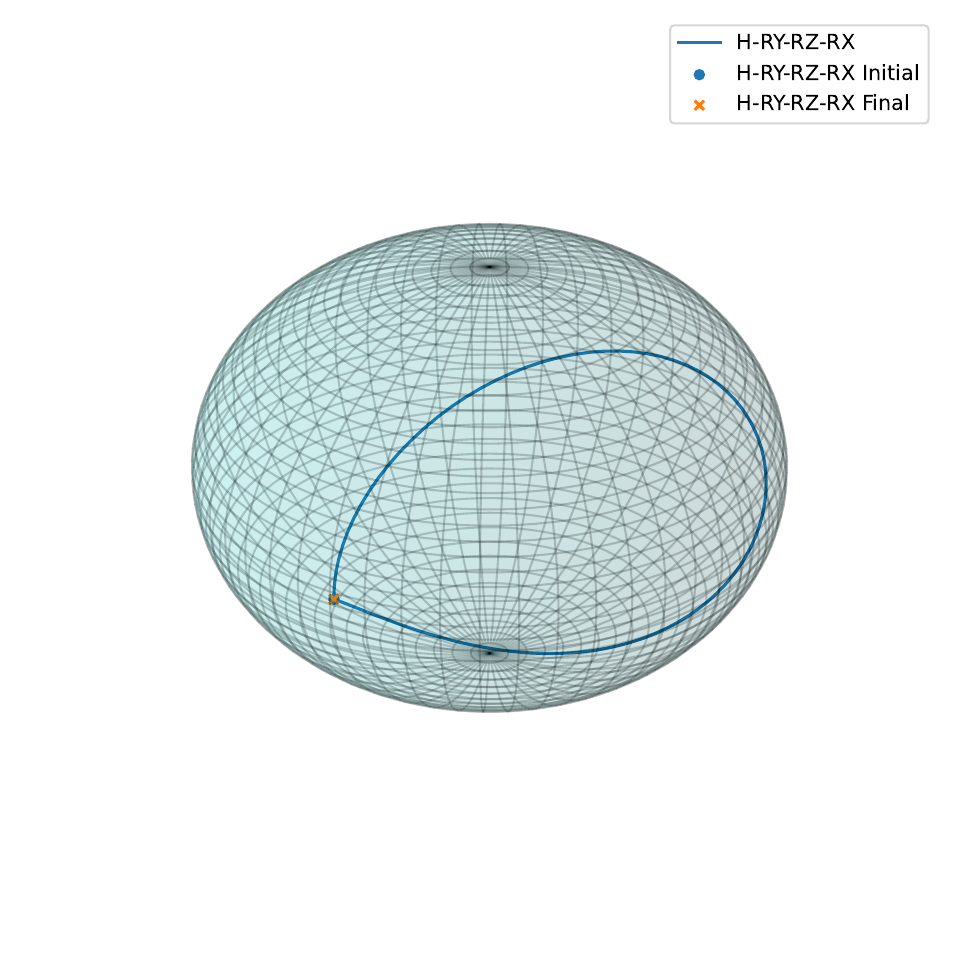}
        \caption{H-RY-RZ-RX}
        \label{subfig:bloch_h_ry_rz_rx}
    \end{subfigure}

\caption{Representations of the trajectories of applying the transformations H-RY, H-RY-RX, H-RY-RX-RZ, H-RY-RZ, and H-RY-RZ-RX varying the angle parameter in the interval [$-\pi / 2$, $\pi / 2$].}
\label{fig:bloch_start_h_ry}
\end{figure*}

\begin{figure*}[h]
    \centering
    \begin{subfigure}[b]{0.3\textwidth}
        \centering
        \includegraphics[width=\textwidth]{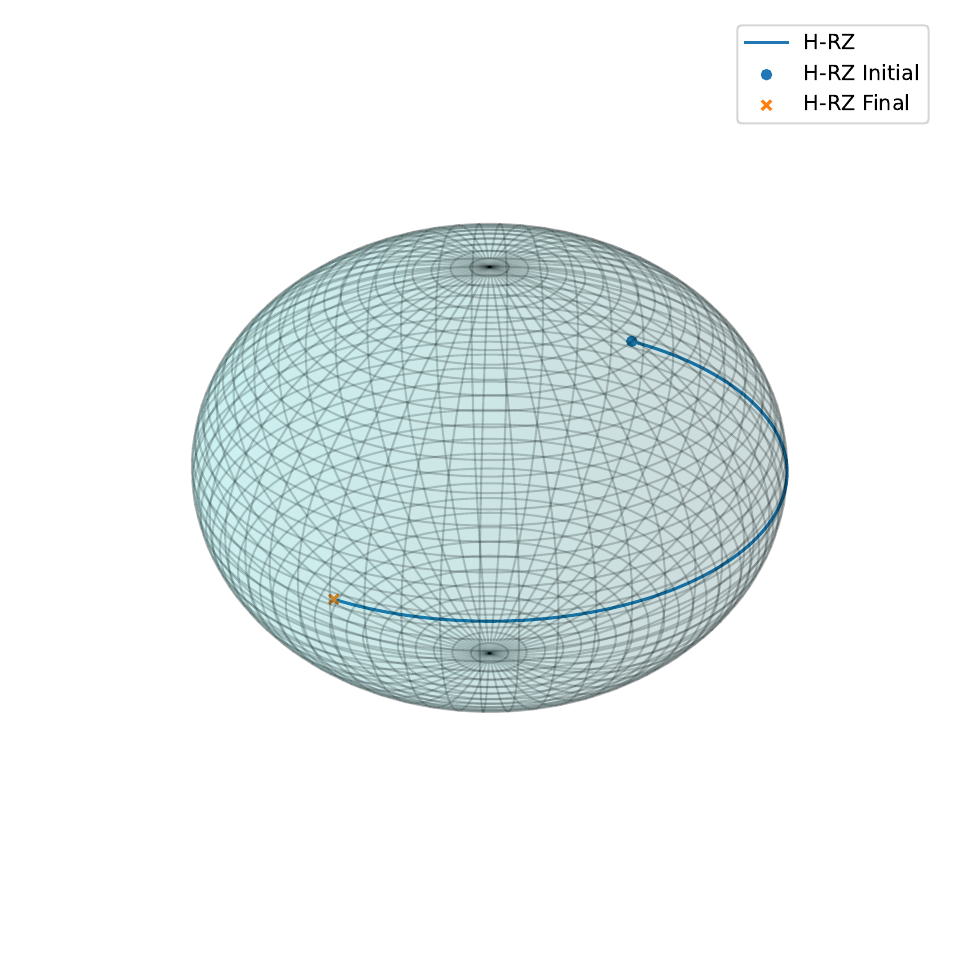}
        \caption{H-RZ}
        \label{subfig:bloch_h_rz}
    \end{subfigure}
    \begin{subfigure}[b]{0.3\textwidth}
        \centering
        \includegraphics[width=\textwidth]{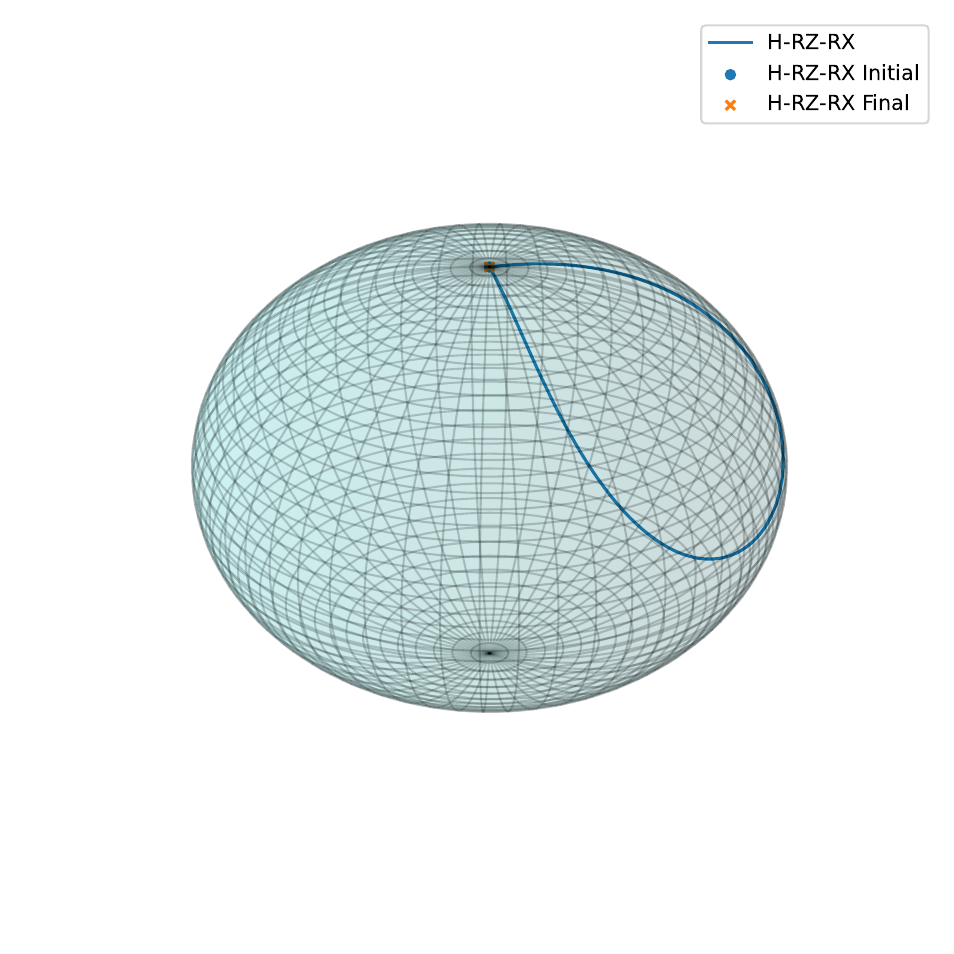}
        \caption{H-RZ-RX}
        \label{subfig:bloch_h_rz_rx}
    \end{subfigure}
    \begin{subfigure}[b]{0.3\textwidth}
        \centering
        \includegraphics[width=\textwidth]{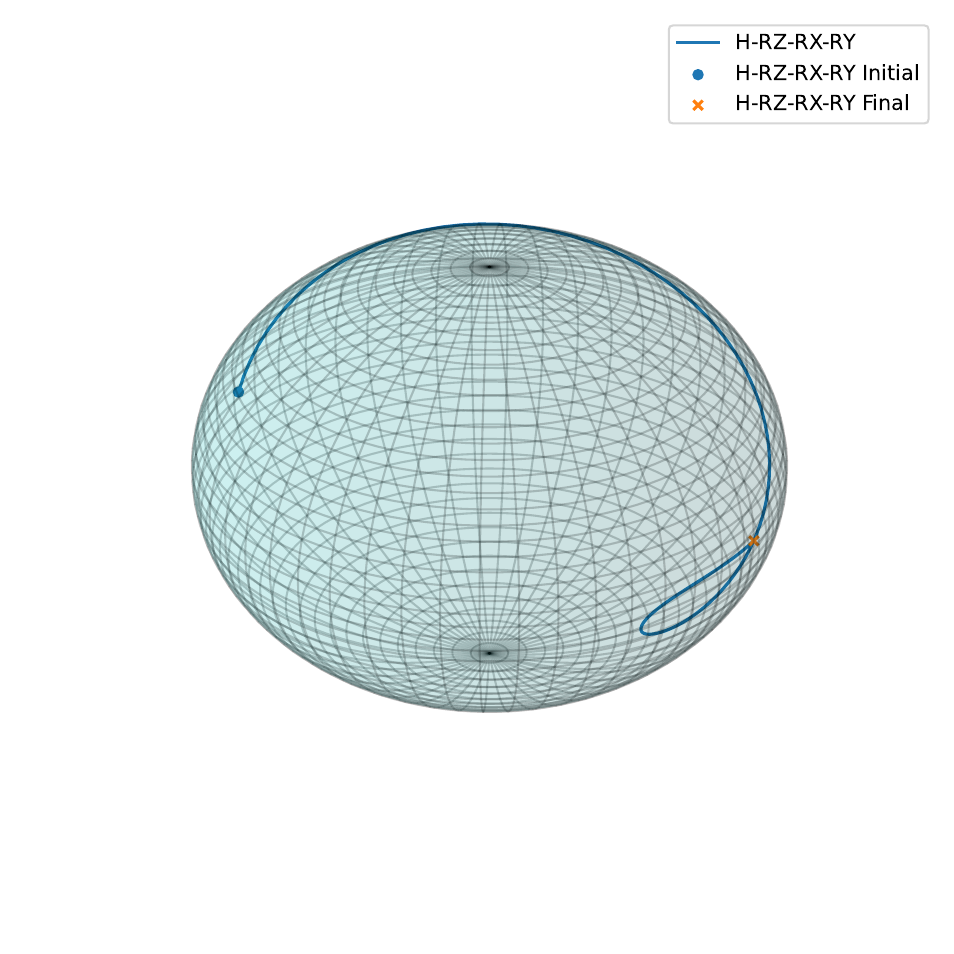}
        \caption{H-RZ-RX-RY}
        \label{subfig:bloch_h_rz_rx_ry}
    \end{subfigure}    
    \begin{subfigure}[b]{0.3\textwidth}
        \centering
        \includegraphics[width=\textwidth]{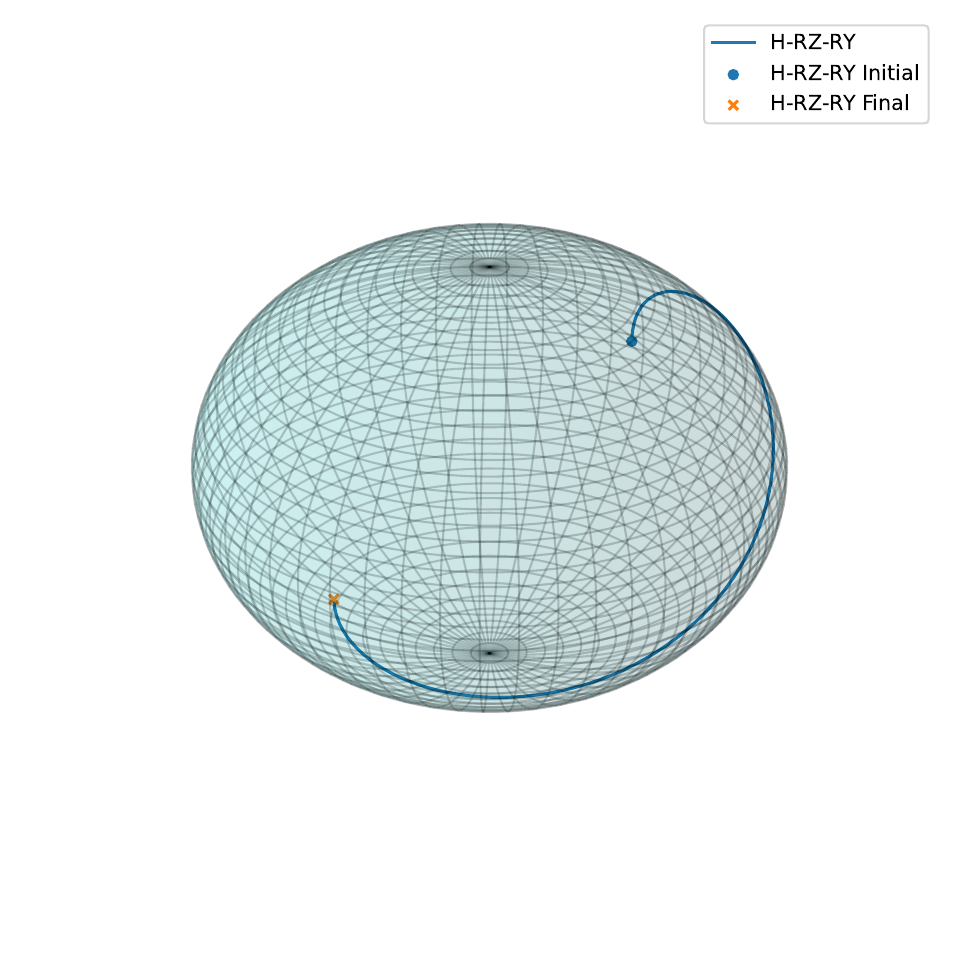}
        \caption{H-RZ-RY}
        \label{subfig:bloch_h_rz_ry}
    \end{subfigure}
    \begin{subfigure}[b]{0.3\textwidth}
        \centering
        \includegraphics[width=\textwidth]{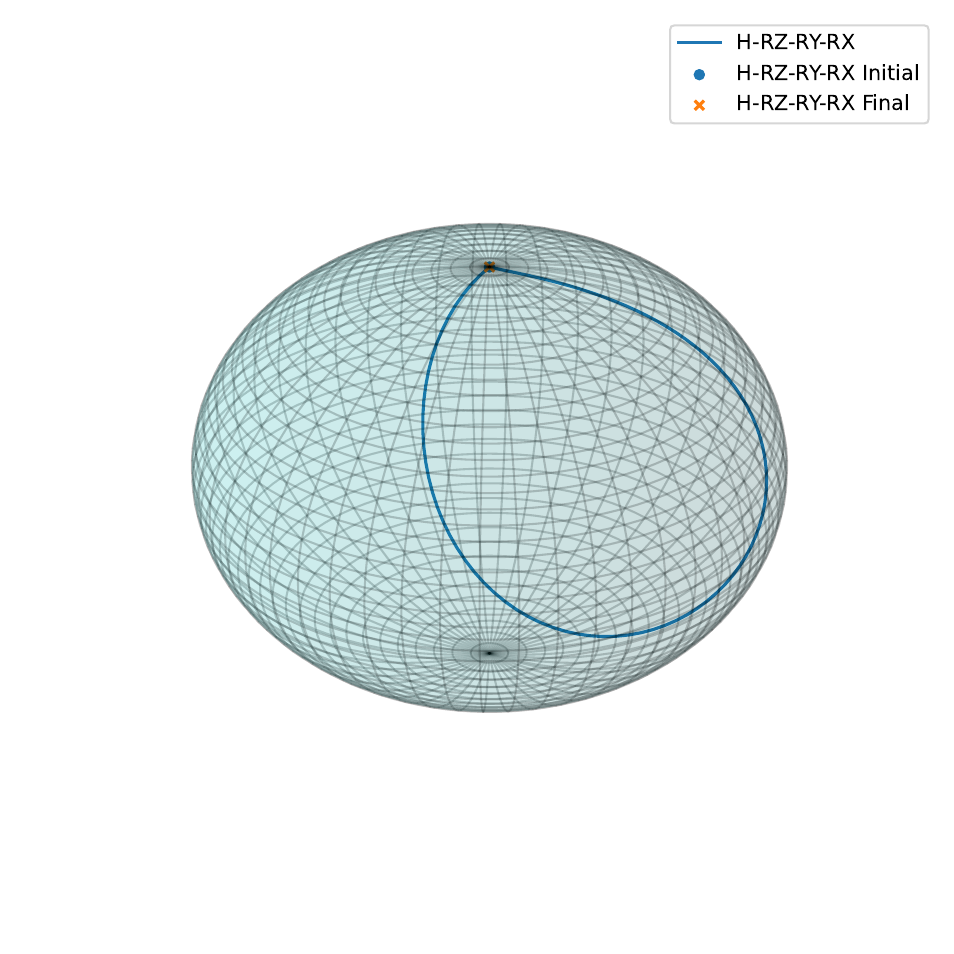}
        \caption{H-RZ-RY-RX}
        \label{subfig:bloch_h_rz_ry_rx}
    \end{subfigure}
    
    \caption{Representations of the trajectories of applying the transformations H-RZ, H-RZ-RX, H-RZ-RX-RY, H-RZ-RY, and H-RZ-RY-RX varying the angle parameter in the interval [$-\pi / 2$, $\pi / 2$].}
\label{fig:bloch_start_h_rz}
\end{figure*}

Other considerations can be made by analyzing the trajectory of the embedding circuits as the input feature varies in the interval
[-$\frac{\pi}{2}$, $\frac{\pi}{2}$], as shown in \autoref{fig:bloch_start_rx}, 
\autoref{fig:bloch_start_ry}, \autoref{fig:bloch_start_h_ry}, and \autoref{fig:bloch_start_h_rz}.

As observed, the complexity of the trajectory increases with the number of applied rotational gates. In the single-rotation case, the trajectory is simply a rotation around the axis associated with that gate. With two rotational gates, the trajectory becomes elliptical, and with three gates, it becomes even more complex.

Adding a Hadamard gate, on the other hand, only alters the starting and ending points of the trajectory. Notably, when using a single-rotation encoding, these points are distant to each other, while in more complex encoding circuits (e.g., those with three rotational gates), they coincide. Consequently, if the features only assume discrete values (e.g. binary features), a single-rotation encoding is preferable. Conversely, for continuous features, by using additional rotational gates may be more suitable.

\subsection{Variational circuit}
In the \gls*{qml} domain, the Ansatz refers to the parametric section of the circuit that allows for the processing of input data. The present study utilizes the Strongly Entangling layer, as proposed by Maria Schuld in \cite{Schuld-basics-2} and implemented as a template in a Pennylane class \cite{StronglyEntanglingLayer}. This circuit comprises two main sections, namely, the one composed by the rotational operators and the other with the entanglement circuit, accomplished via CNOT gates. \autoref{fig:VQCLoop} depicts an example of an ansatz of three layers. \\

The rotational operators involve the use of ROT gates, which are defined by the parameters $\phi$, $\theta$, and $\gamma$, and their transformation matrix is given by \autoref{eq:ROT-gate}. It is important to highlight that a generic unitary transformation can be accomplished by employing such gates.

\begin{equation}
\label{eq:ROT-gate}
\begin{split}
Rot(\phi, \theta, \gamma) = RZ(\phi)RY(\theta)RZ(\gamma) = \\ =
\begin{pmatrix}
e^{-i(\phi + \gamma)/2} \cos(\theta / 2) & -e^{i(\phi - \gamma)/2} \sin (\theta / 2) \\
e^{-i(\phi - \gamma)/2} \sin (\theta / 2) & e^{i(\phi + \gamma)/2} \cos (\theta / 2) \\
\end{pmatrix}
\end{split}
\end{equation}

The main goal of the Entanglement Circuit shown is to create significant entanglement between qubits. The placement of CNOT gates guarantees that the qubits are highly correlated allowing a larger portion of the Hilbert space to be explored efficiently, making the quantum algorithm more powerful. Furthermore, this entanglement circuit has demonstrated greater expressibility, as evidenced in \cite{https://doi.org/10.1002/qute.201900070}.

One of the circuit parameters is the number of layers that constitutes the ansatz. To reduce computational costs associated with finding the optimal number, models were tested on a limited set of layers, progressing in multiples of 2 (2, 4, 6, 8, 10).

\subsection{Re-uploading technique}
The present work studies the effects of an additional technique introduced in \autoref{sec:Reup}, the ``re-uploading", which involves embedding data into the quantum circuit each time a new ansatz is inserted. This method is utilized by certain models employed in the study, specifically those that leverage Angle or Amplitude encoding. The performance of these models is compared to that of models not implementing this technique. 

\subsection{Testing methodology}
Once the model has been defined, it is essential to consider the entire training and testing phase, beginning with data preparation. The data must be first normalized between 0 and $\pi$. Then, to ensure that both angle-encoded and amplitude-encoded models have the same number of qubits, Principal Component Analysis (PCA)\cite{wold1987principal} is applied to the angle-encoded models. This reduces the number of features from $N$ to $\log_2 N$.\\

This work employs real datasets to benchmark different encoding strategies. They are divided into training and test sets, which comprise 80\% and 20\% of the initial dataset, respectively.\\

Following dataset preparation, the training phase can begin. Each parameter is initialized at a random value and then optimized by exploiting the Adam algorithm \cite{kingma2014adam}. This is a classical stochastic gradient descent optimization mechanism with an adaptive learning rate. In particular, the initial learning rate value has been set to 0.01, which represents a trade-off between a too-large value for which the optimizer can diverge from the optimal solution and a too-small one for which a higher number of steps is required.\\
The training process has been repeated for 30 epochs, i.e., considering 30 times the overall training dataset. Furthermore, the mini-batch strategy has been applied, where the dataset has been divided into subsets of 10 elements, and the parameters have been updated only when the evaluation of the subset has been completed. The training process was stopped at this limited number of epochs because we observed that the variations in both loss and accuracy during the final epochs were minimal.\\
Each model has been trained and tested ten times to ensure a fair comparison. Therefore, the results shown in the successive sections represent mean values.

\subsection{Settings}
All the quantum circuits considered in this study have been developed using the Pennylane \cite{bergholm2022pennylane} library \textit{(version 0.33.0)}, an open-source library for defining quantum machine learning, quantum chemistry, and quantum computing applications. The circuits have been simulated using the standard Pennylane qubit-based device, called \textit{``default.qubit"}, with a default value of 1000 shots. This is an ideal simulator for testing the functionality of QML models. For the device, it's important to specify the desired number of qubits, which is provided as a parameter in the class definition. In our case, the number of qubits is set to four for the models applied to the Wine dataset, and three for those used with the Diabetes dataset.\\
In addition, the circuits have been interfaced with the PyTorch \cite{paszke2017automatic} library for the training phase to enhance the model's performance further. This integration has enabled classical parameter back-propagation. This approach has been preferred over the parameter-shift rule to reduce the implementation time. Because of its integration with PyTorch, we used PennyLane rather than Qiskit for implementing our quantum models.

\section{Results}

The models were tested on two distinct datasets: the Wine \cite{misc_wine_109} dataset and the Diabetes \cite{diabetes_dataset}  dataset. The Wine dataset consists of 13 features and 178 samples, each classified into one of three classes. It's important to note that this dataset is unbalanced, with 59 samples in the first category, 71 in the second, and 48 in the third. On the other hand, the Diabetes dataset represents a binary classification problem with 168 items. Each one is characterized by eight features, including Pregnancy, Glucose, Blood Pressure, Skin Thickness, Insulin, BMI, Diabetes Pedigree Function, and Age. Each sample is classified as either class 0 (non-diabetic) or class 1 (diabetic). Also this dataset is unbalanced, with 65\% of the data belonging to class 0 and the remaining 35\% to class 1.

For the Wine dataset, model performance is evaluated using accuracy as the primary metric. However, given the stronger imbalance in the Diabetes dataset, additional metrics are needed to evaluate the model’s effectiveness. 

\begin{figure}
    \centering
    \includegraphics[width=0.7\linewidth]{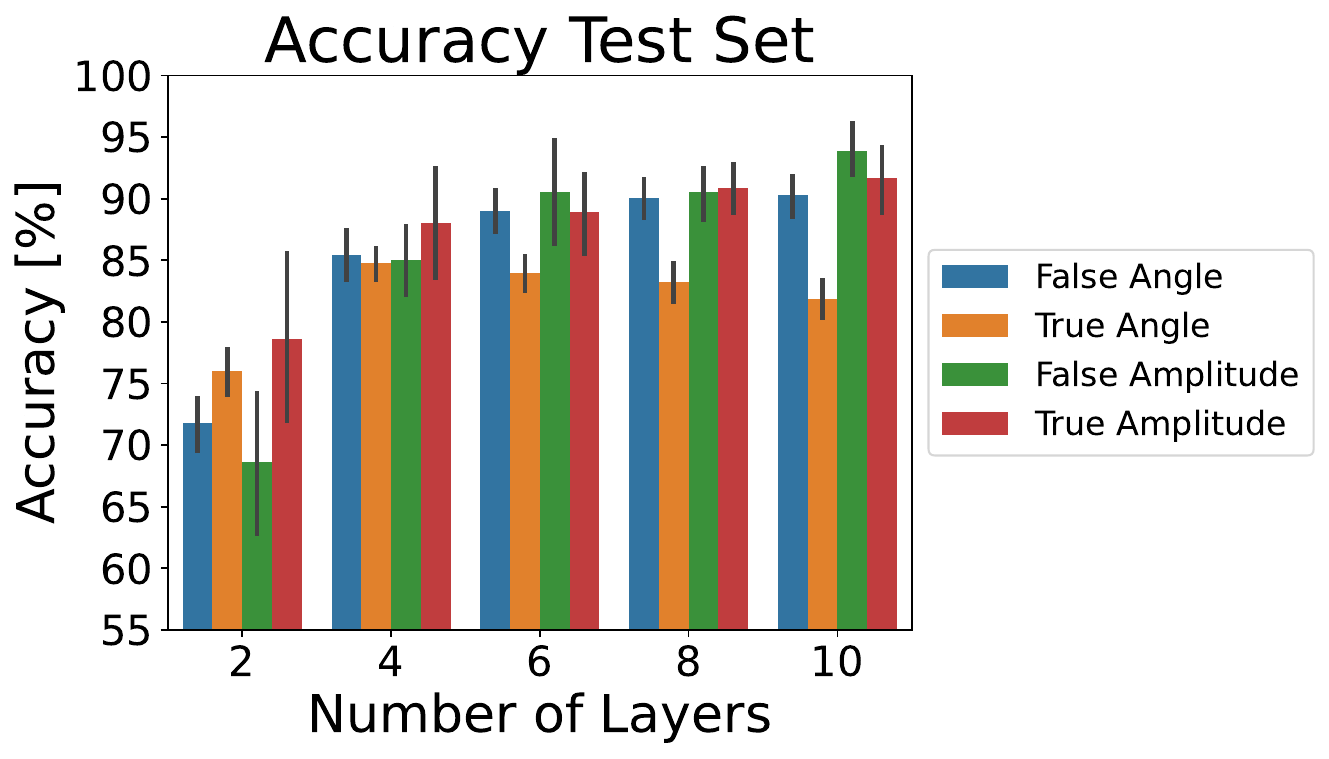}
    \caption{Comparison of models utilizing and not utilizing the re-uploading technique, along with Angle and Amplitude encoding methods, in terms of accuracy on the Diabetes dataset. The results show that the Amplitude encoding models have obtained higher performance for the VQC with more than 2 layers. Moreover, the re-uploading strategy does not guarantee higher accuracy results in this case.}
    \label{fig:Wine-Accuracy}
\end{figure}

Regarding the Wine dataset, models using Amplitude encoding methods have, on average, obtained better accuracy compared to those using Angle encoding, as illustrated in Figure \ref{fig:Wine-Accuracy}. Despite this, the highest-performing model in this task employed Angle encoding with an RY gate for embedding, with 10 Strongly Entangling layers and no re-uploading. The encoding circuit was found to have a substantial impact on accuracy. In fact, the difference in accuracy between the best and worst model, holding the number of layers constant and using re-uploading, averaged 33\%, with a maximum difference of 41.1\%. The results are shown in \autoref{tab:Acc_Wine_Comp}, reporting the accuracy percentage for both the best and the worst model, as well as the associated standard deviation. In the same table, the accuracy difference between the best and worst models and the associated standard deviation ($\sqrt{STD_{best}^2 + STD_{worst}^2}$) are also calculated.

For the Diabetes classification task, similar observations were made regarding the influence of the encoding strategy on model performance. When comparing various performance metrics, a significant disparity emerged between the best and worst models. These results were analyzed using accuracy (reported in \autoref{tab:Acc_Diabetes_Comp}), balanced accuracy (\autoref{tab:Acc_Wine_Comp_BalancedAcc}), recall (\autoref{tab:Acc_Wine_Comp_Recall}), precision (\autoref{tab:Acc_Wine_Comp_Precision}), and F1-score (\autoref{tab:Acc_Wine_Comp_F1}).
For each of the tables considered, not only the results of the metrics are provided, but also their associated standard deviation values. In addition, the difference between the best and worst model for each of the metrics considered and the associated standard deviation are calculated as well.

In terms of accuracy, the average difference between the best and worst models with identical architectures (same re-uploading process and number of layers) was approximately 8\%, with the maximum difference reaching 10.3\%. This gap became even more pronounced when considering balanced accuracy, where the average difference increased to 12\%, with a maximum of 19\%. Additionally, differences in precision, recall, and F1-score between the best and worst models remained significant, emphasizing the importance of encoding method optimization. However, these metrics exhibited greater uncertainties, which can depend on the intrinsic characteristics of the dataset or of the model. Further investigation on these results can be done to understand on what depends this uncertainty.

Another analysis addressed, models were compared based on the encoding strategy they employed—grouped by Angle and Amplitude encoding—and the use of the re-uploading technique across all considered metrics (Accuracy, Balanced Accuracy, Recall, Precision, and F1-score). The models that exploit Angle encoding consistently outperformed their counterparts with Amplitude encoding on all metrics. Additionally, models that employed the re-uploading strategy showed an average improvement in classification performance, particularly on Balanced Accuracy, Recall, and F1-score, compared to models that did not use it.
The selection of the best model depends on the specific classification goal. If balanced accuracy and F1-score are prioritized, the best model is the RX-RZ-RY with 10 layers and no re-uploading. If precision is the primary focus, the RX-RY-RZ with 10 layers and no re-uploading is the optimal choice. For maximizing recall, the RX-RZ-RY with 6 layers and no re-uploading performs best.

All the collected results highlight the critical impact of encoding methods on model classification performance, reinforcing the idea that encoding should be treated as a key hyper-parameter that must be carefully optimized to enhance both model accuracy and efficiency. Refining the encoding process can lead to significant improvements in the model’s overall predictive capabilities.\\

\begin{figure*}
    \centering
         \begin{subfigure}[b]{0.32\textwidth}
         \centering
         \includegraphics[width=\linewidth]{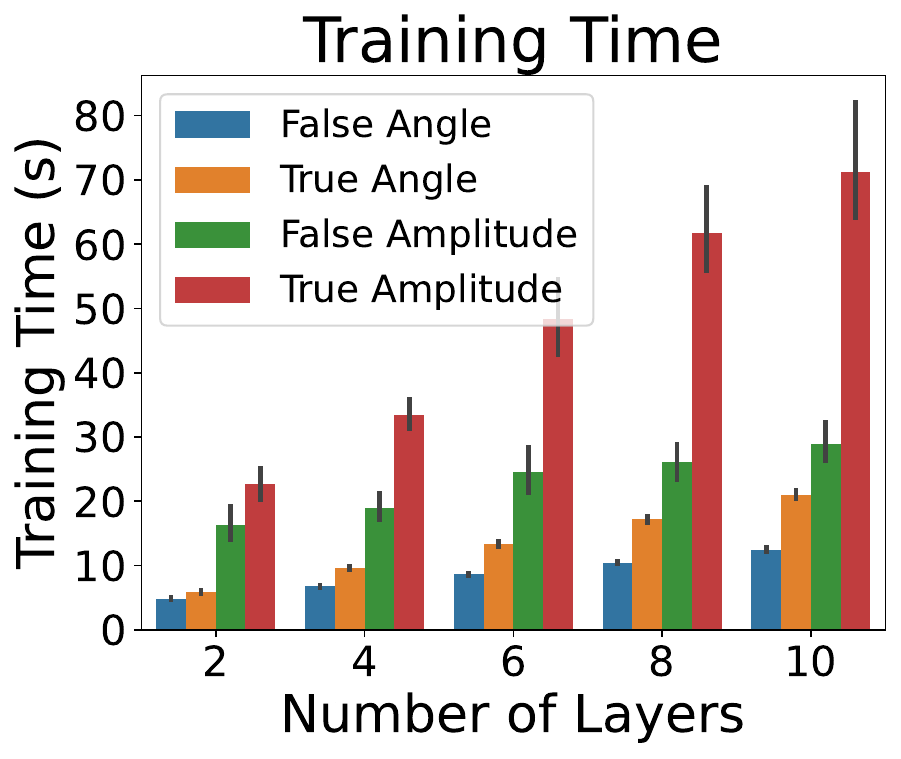}
         \caption{}
         \label{fig:TrainingTime_wine}
         \end{subfigure}
         \hfill         
         \begin{subfigure}[b]{0.32\textwidth}
         \centering
         \includegraphics[width=\linewidth]{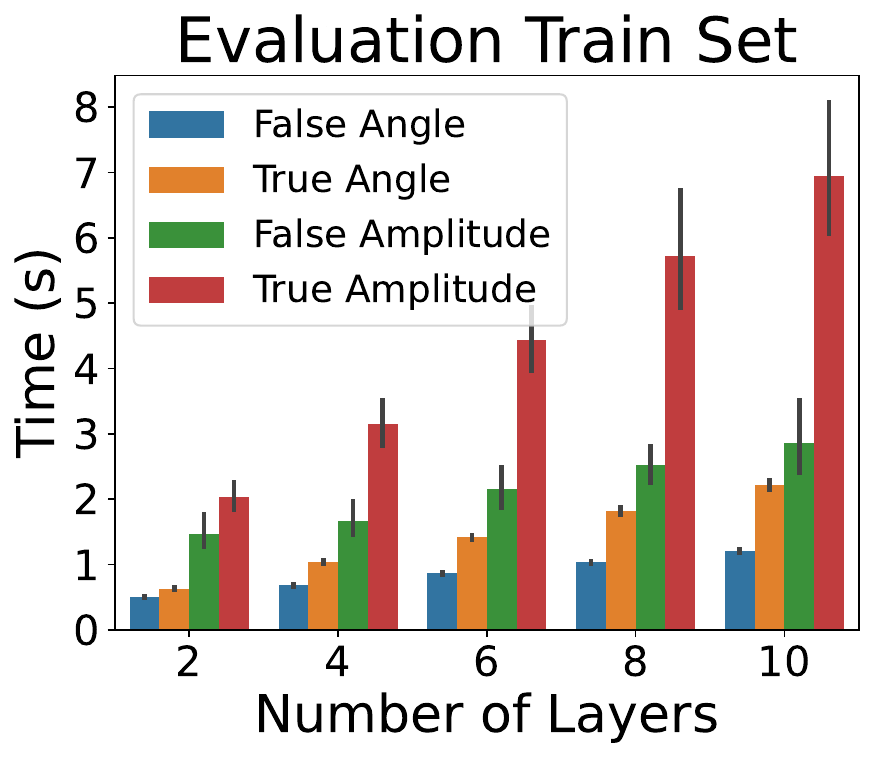}
         \caption{}
         \label{fig:EvalTrainTime_wine}
         \end{subfigure}
         \hfill
         \begin{subfigure}[b]{0.32\textwidth}
         \centering
         \includegraphics[width=\linewidth]{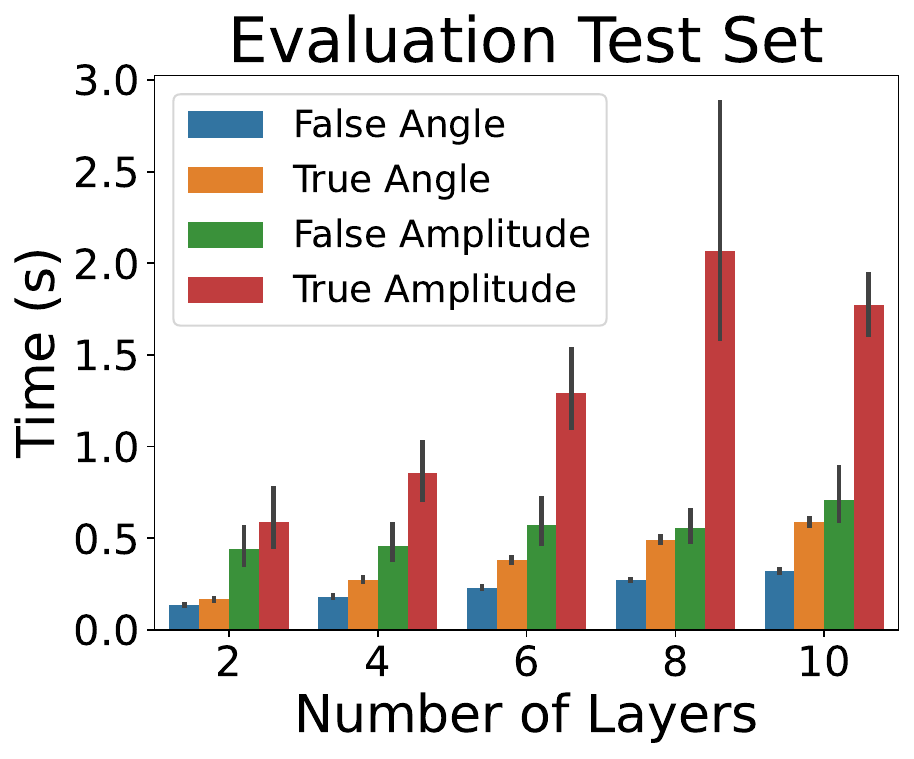}
         \caption{}
         \label{fig:EvalTestTime_wine}
         \end{subfigure}
         \begin{subfigure}[b]{0.32\textwidth}
         \centering
         \includegraphics[width=\linewidth]{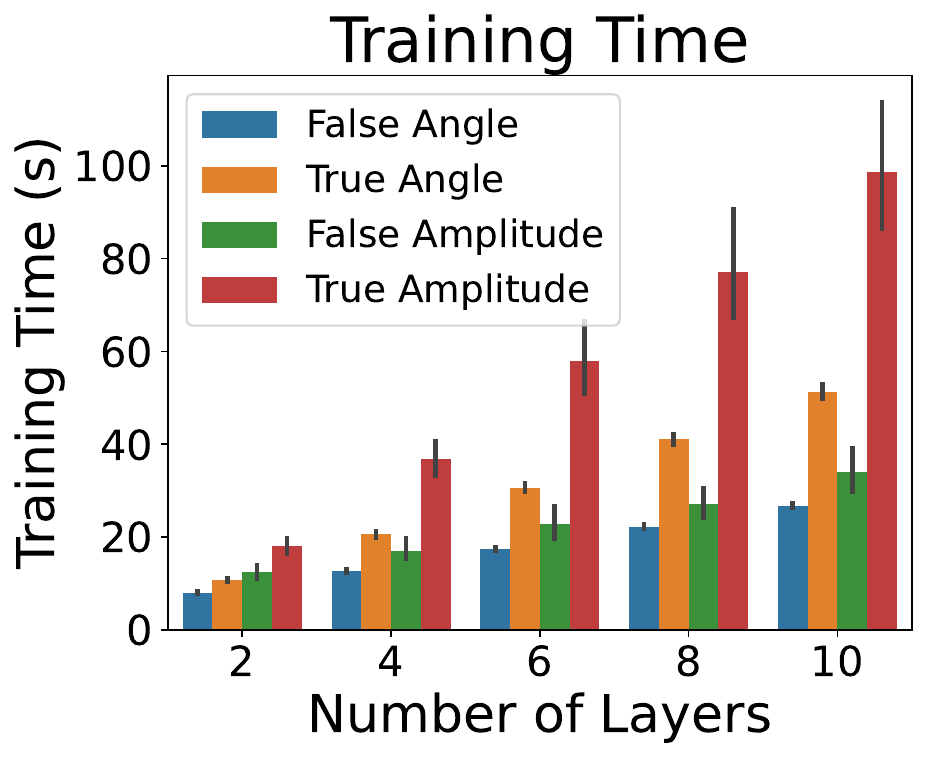}
         \caption{}
         \label{fig:TrainingTime_diabetes}
         \end{subfigure}
         \hfill         
         \begin{subfigure}[b]{0.32\textwidth}
         \centering
         \includegraphics[width=\linewidth]{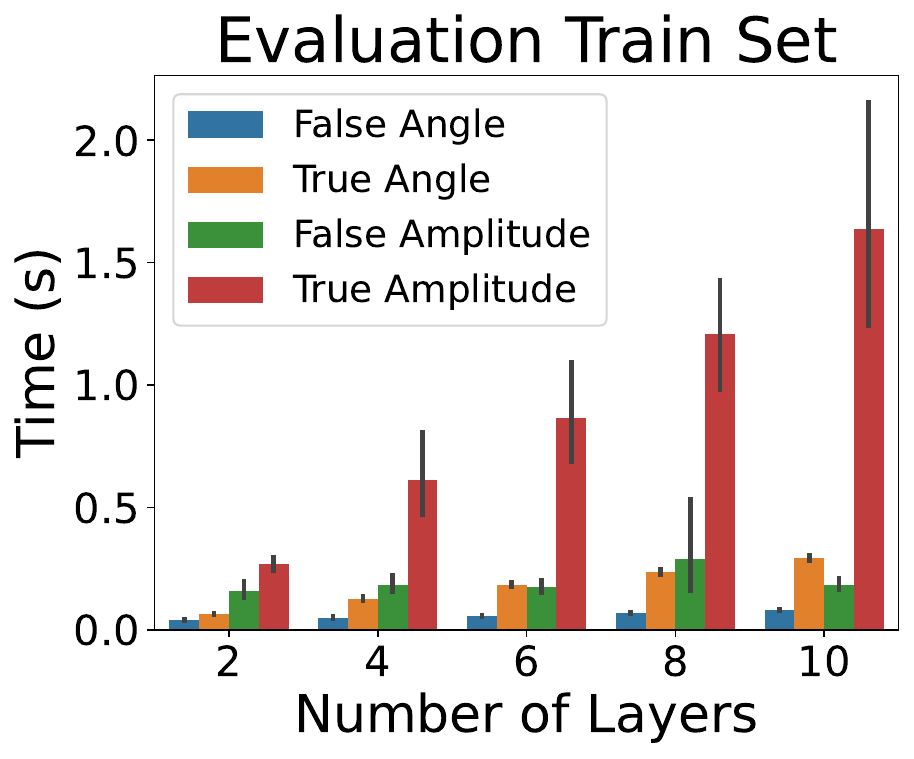}
         \caption{}
         \label{fig:EvalTrainTime_diabetes}
         \end{subfigure}
         \hfill
         \begin{subfigure}[b]{0.32\textwidth}
         \centering
         \includegraphics[width=\linewidth]{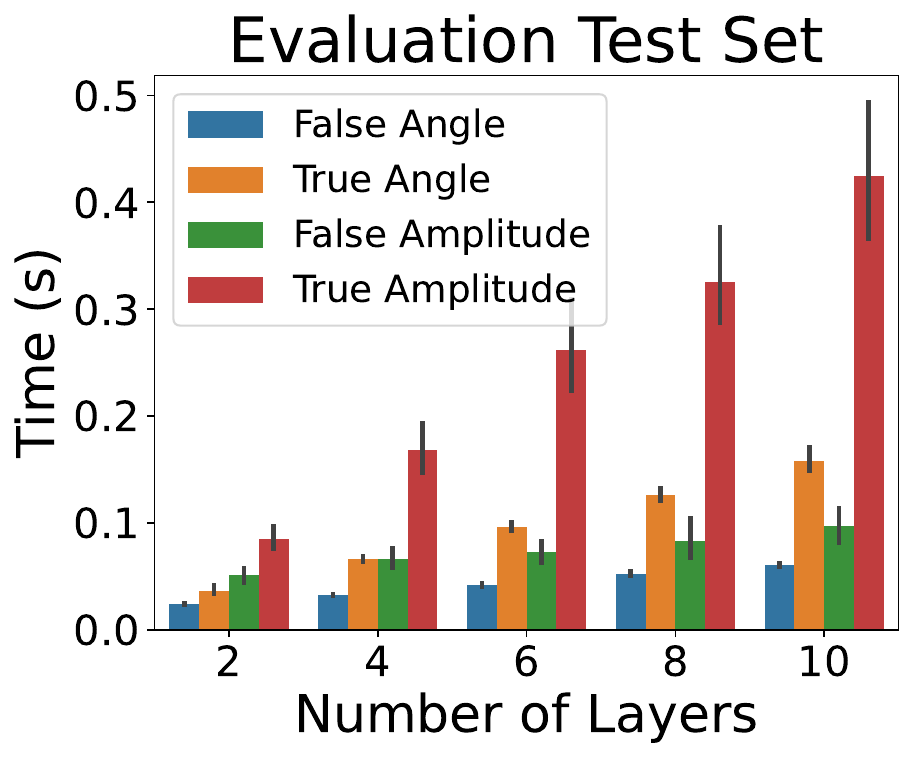}
         \caption{}
         \label{fig:EvalTestTime_diabetes}
         \end{subfigure}
         
         \caption{Illustration of the time required to train the model (\autoref{fig:TrainingTime_wine}), to evaluate the training set (\autoref{fig:EvalTrainTime_wine}), and to evaluate the test set (\autoref{fig:EvalTestTime_wine}) on the Wine dataset and on the Diabetes dataset (respectively \autoref{fig:TrainingTime_diabetes}, \autoref{fig:EvalTrainTime_diabetes}, \autoref{fig:EvalTestTime_diabetes}). The four bars displayed represent: models utilizing Angle encoding without reuploading, models employing the reuploading technique with Angle encoding, models using Amplitude encoding without reuploading, and models with both Amplitude encoding and reuploading. Results indicate on both analyses that the Amplitude encoding circuit is the slowest. Furthermore, as the number of layers increases, so does the time needed for model training and inference. Finally, the application of the reuploading technique involves additional training and inference time.}
         \label{fig:TimeWine}
\end{figure*}

Moreover, the study examines the effect of encoding strategies on model simulation times for both the Wine and Diabetes datasets, as shown in \autoref{fig:TimeWine}. The results were obtained through simulation in \cite{processorservervlsi}, with the number of threads constrained to 24. These results depend on system parameters, such as the number of CPU cores, CPU frequency, and the number of threads, among others. Additionally, since the simulations were not conducted on a real device, it is not possible to guarantee that the measured times directly correlate with the times required on actual hardware. Training and inference times for both the training and test sets are reported. On average, models employing Amplitude encoding (using the Mottonen circuit) took significantly longer to train and infer than those using Angle encoding. As expected, models utilizing the re-uploading strategy also required more time for both training and inference.

\label{sect:wine}

\begin{figure*}
    \centering
    \begin{subfigure}[b]{0.45\textwidth}
         \centering
         \includegraphics[width=\linewidth]{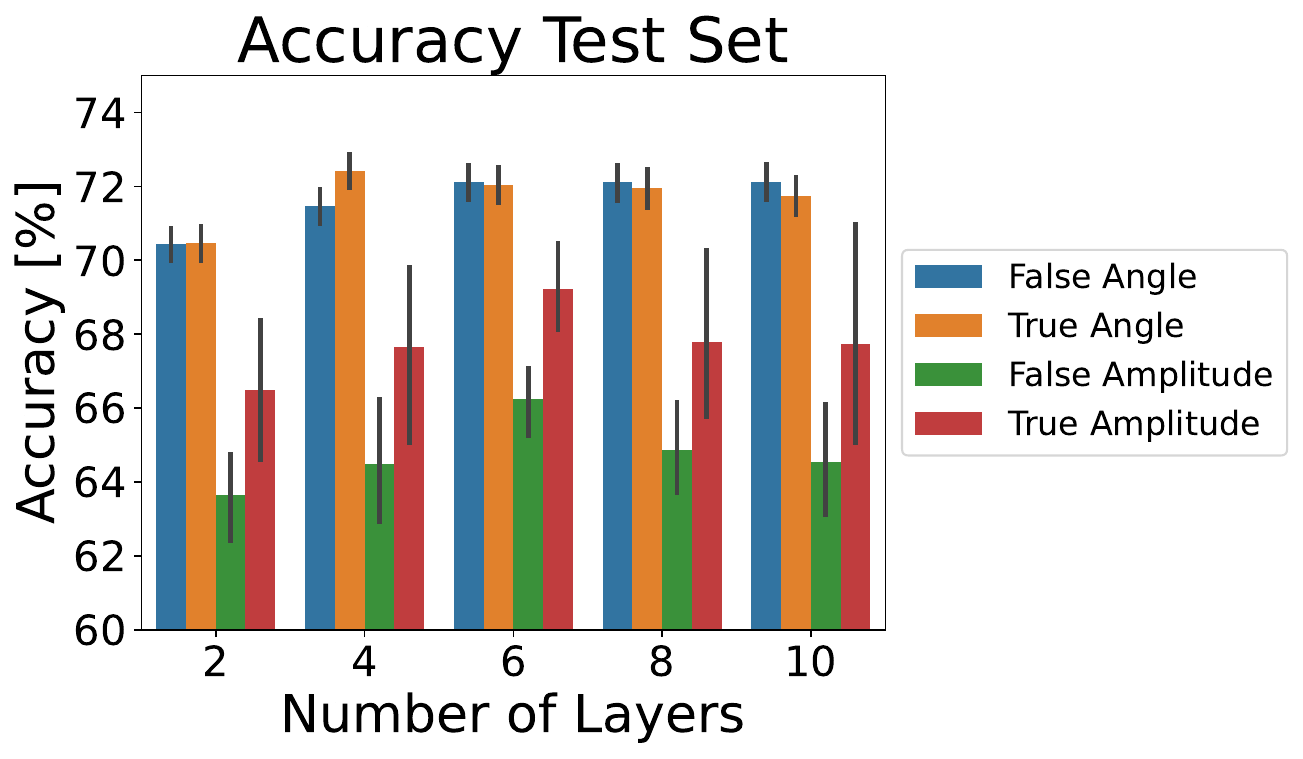}
         \caption{}
         \label{fig:Diabetes-Accuracy}
     \end{subfigure}
    \hfill
    \begin{subfigure}[b]{0.45\textwidth}
         \centering
         \includegraphics[width=\linewidth]{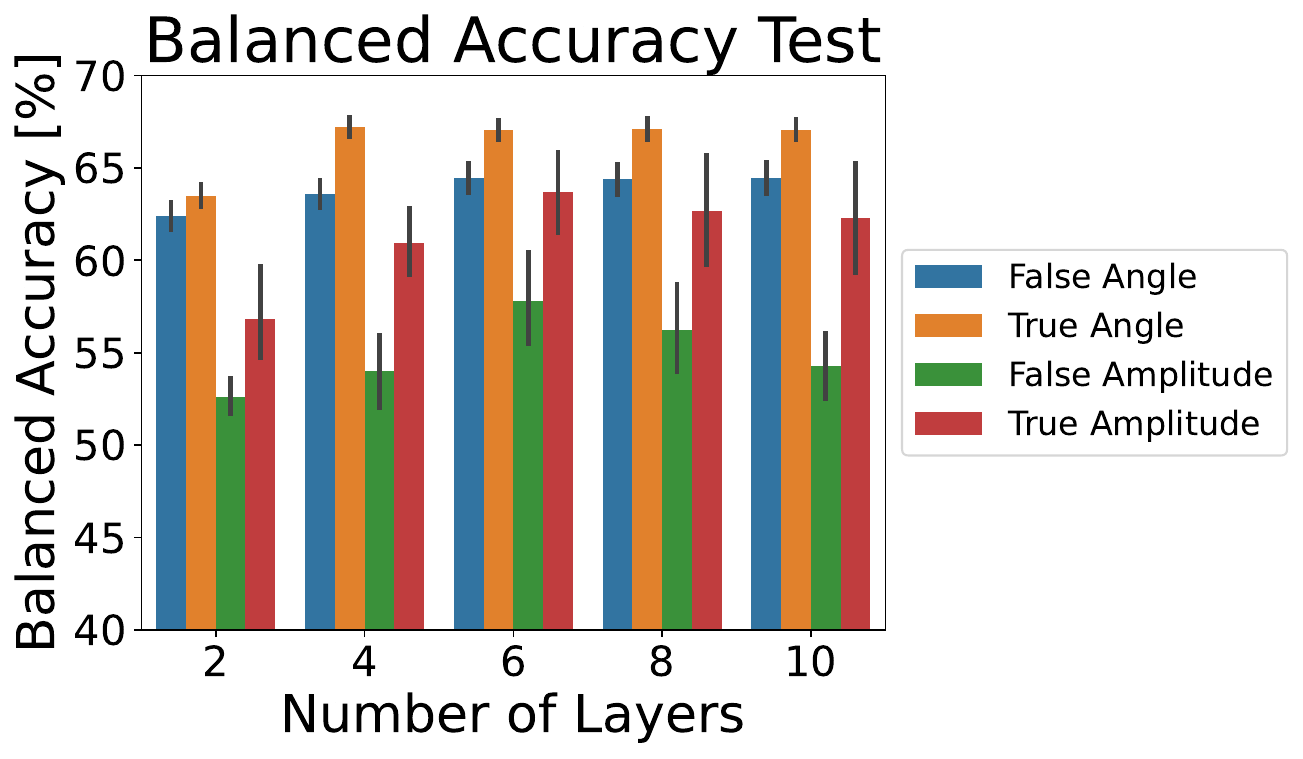}
         \caption{}
         \label{fig:Diabetes-Balanced-Accuracy}
     \end{subfigure}
     \hfill
    \begin{subfigure}[b]{0.45\textwidth}
         \centering
         \includegraphics[width=\linewidth]{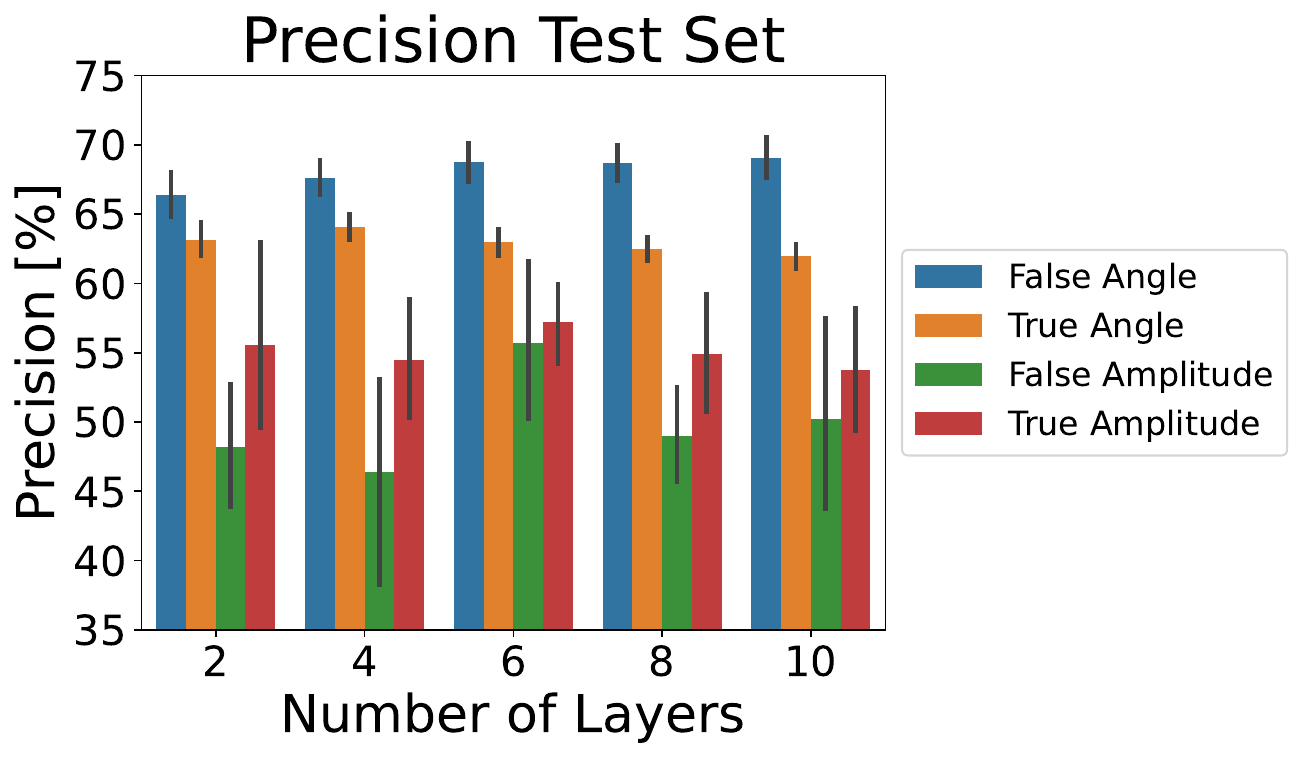}
         \caption{}
         \label{fig:Diabetes-Precision}
     \end{subfigure}
     \hfill
    \begin{subfigure}[b]{0.45\textwidth}
         \centering
         \includegraphics[width=\linewidth]{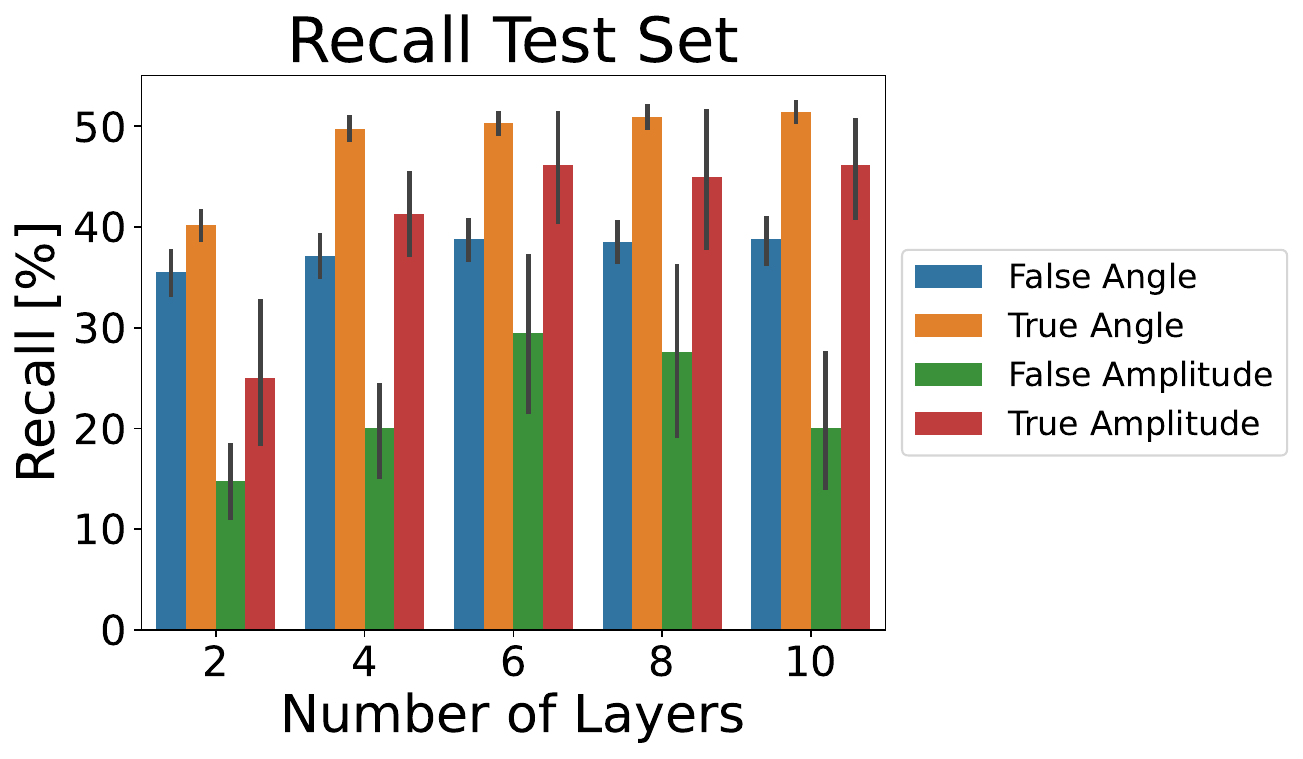}
         \caption{}
         \label{fig:Diabetes-Recall}
     \end{subfigure}
     \hfill     
    \begin{subfigure}[b]{0.45\textwidth}
         \centering
         \includegraphics[width=\linewidth]{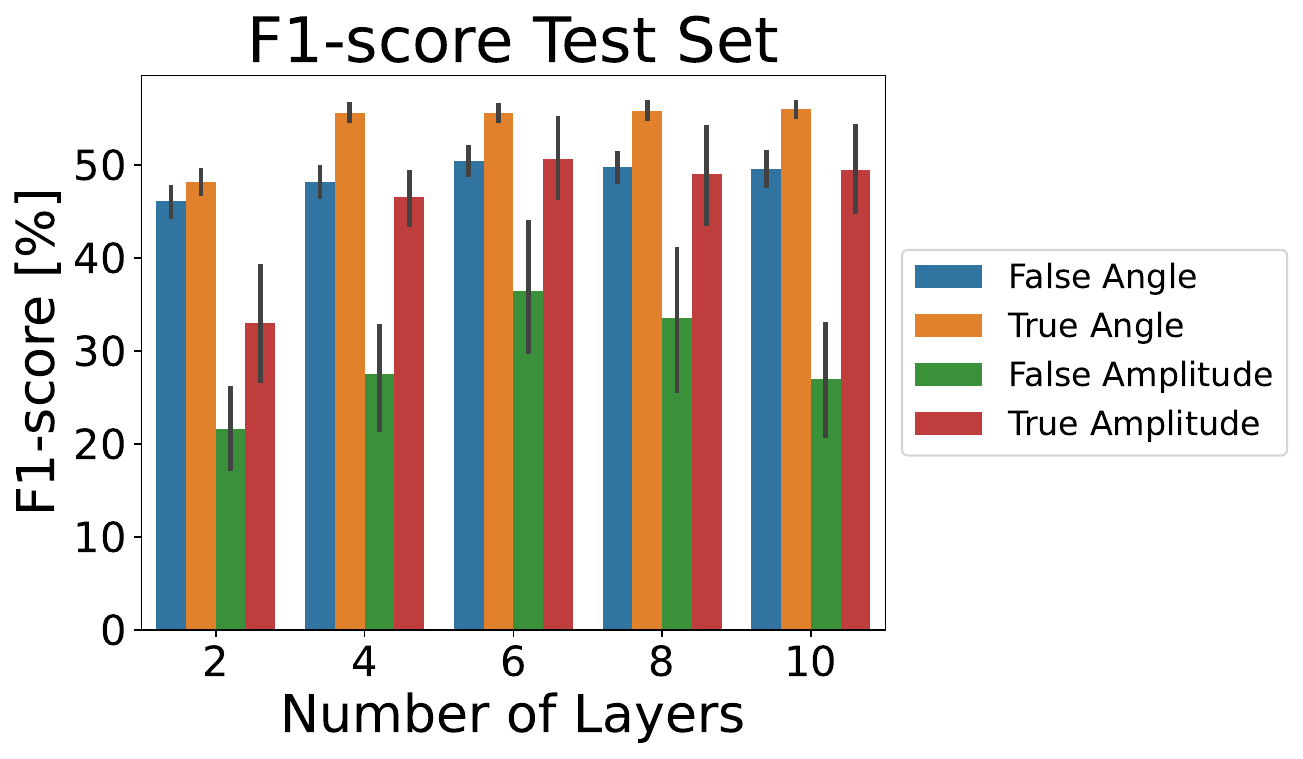}
         \caption{}
         \label{fig:Diabetes-F1}
     \end{subfigure}

    \caption{Comparison of models utilizing and not utilizing the re-uploading technique, along with Angle and Amplitude encoding methods, in terms of Accuracy (\autoref{fig:Diabetes-Accuracy}), Balanced Accuracy (\autoref{fig:Diabetes-Balanced-Accuracy}), Precision (\autoref{fig:Diabetes-Precision}), Recall (\autoref{fig:Diabetes-Recall}), and F1-score (\autoref{fig:Diabetes-F1}). The results indicate that Angle encoding models generally outperform others across all the metrics evaluated. Furthermore, models utilizing the re-uploading technique demonstrate superior performance in terms of Balanced Accuracy, Recall, and F1-score for both Angle and Amplitude encoding. However, for Accuracy and Precision, the re-uploading technique provides improvements exclusively for Amplitude encoding models.}
    \label{fig:Diabetes}
\end{figure*}

\begin{table*}[h!]
  \centering
  \small
  \resizebox{\textwidth}{!}{%
  \begin{tabular}{cc|cccccc|cc}
    \toprule
    &&\multicolumn{3}{c}{\textbf{Best Results}} & \multicolumn{3}{c|}{\textbf{Worst Results}}& & \\
\textit{Layers}&\textit{RU}&\textit{Encoding}&\textit{Accuracy (\%)}&\textit{STD (\%)}&\textit{Encoding}&\textit{Accuracy (\%)}&\textit{STD (\%)}&\textit{$\Delta_{Acc.}$(\%)}&\textit{$\Delta_{Acc}$ STD (\%)}  \\ \hline 
         2 & False &H-RZ&90.000000 &6.029193 &H-RY-RZ-RX&54.722222 &9.170874 &35.277778 &10.975250 \\
         2 & True &RX&86.944444 &6.421682 &RX-RY-RZ&62.222222 &13.993434 &24.722222 &15.396564 \\
         4 & False&RX&95.000000&4.498133 &RX-RY-RZ&51.111111 &6.441677 &43.888889 &7.856742 \\
         4 & True &H-RY&93.611111 &3.220838 &H-RZ-RY-RX&71.111111 &6.307180 &22.500000 &7.081972 \\
         6 &False &RY&97.222222 &3.207501 &RX-RY-RZ&56.388889 &3.939268 &40.833333 &5.079951 \\
         6 &True &RY&93.888889 &5.204989 &RX-RZ-RY&64.722222&8.788983&29.166667 &10.214604 \\
         8 &False &H-RZ&96.944444 &2.762303 & RX-RY-RZ&59.166667 &9.537487 &37.777778 &9.929449 \\
         8 &True &H-RZ&91.388889 &3.574122 &RY-RX-RZ&63.611111 &8.924504 &27.777778 &9.613591 \\
        10 &False &RY&97.500000 &2.432208 &RX-RY-RZ&56.388889 &5.406963 &41.111111 &5.928819 \\
        10 &True &H-RZ&92.500000 &4.545532 &H-RY-RX-RZ&61.388889 &9.925131 &31.111111 &10.916506 \\
\bottomrule
\end{tabular}
  }
  \caption{Comparison between the best and the worst encoding models grouped by the number of parametric layers and use of the re-uploading (RU) technique for classification on the Wine dataset. On average, the variation in terms of test accuracy is around 33\%.}
  \label{tab:Acc_Wine_Comp}
\end{table*}

\begin{table*}[h!]
  \centering
  \small
  \resizebox{\textwidth}{!}{%
  \begin{tabular}{cc|cccccc|cc}
    \toprule
    &&\multicolumn{3}{c}{\textbf{Best Results}} & \multicolumn{3}{c|}{\textbf{Worst Results}}& & \\
\textit{RU}&\textit{Layers}&\textit{Encoding}&\textit{Accuracy (\%)}&\textit{STD (\%)}&\textit{Encoding}&\textit{Accuracy (\%)}&\textit{STD (\%)}&\textit{$\Delta_{Acc.}$(\%)}&\textit{$\Delta_{Acc}$ STD (\%)}  \\ \hline 
       2 &        False &           H-RZ-RY-RX &       72.402597 &       3.064892 &      Amplitude &        63.636364 &        1.960040 &                 8.766234 &                3.638038 \\
         2 &         True &             RX &       72.987013 &       2.689563 &      Amplitude &        66.493506 &        3.184100 &                 6.493506 &                4.168002 \\
         4 &        False &            H-RY-RX &       73.961039 &       3.246032 &      Amplitude &        64.480519 &        2.855995 &                 9.480519 &                4.323590 \\
         4 &         True &            H-RY-RZ &       74.610390 &       3.387290 &      Amplitude &        67.662338 &        4.023879 &                 6.948052 &                5.259785 \\
         6 &        False &            RX-RZ &       74.350649 &       3.286196 &            H-RY-RZ-RX &        65.194805 &        0.698031 &                 9.155844 &                3.359514 \\
         6 &         True &             H-RY &       73.961039 &       2.379968 &            RY-RX-RZ &        67.857143 &        3.982331 &                 6.103896 &                4.639311 \\
         8 &        False &           RX-RZ-RY &       74.285714 &       3.425115 &      Amplitude &        64.870130 &        2.237932 &                 9.415584 &                4.091424 \\
         8 &         True &            RY-RX &       74.155844 &       2.785404 &            H-RY-RX-RZ &        67.727273 &        3.225762 &                 6.428571 &                4.261926 \\
        10 &        False &           RX-RZ-RY &       74.935065 &       2.471706 &      Amplitude &        64.545455 &        2.582932 &                10.389610 &                3.575034 \\
        10 &         True &             RY &       74.025974 &       2.505592 &            H-RZ-RY-RX &        67.662338 &        2.575667 &                 6.363636 &                3.593334 \\
\bottomrule
\end{tabular}
  }
  \caption{Comparison between the best and the worst encoding models in terms of accuracy grouped by the number of parametric layers and use of the re-uploading (RU) technique for classification on the Diabetes dataset. On average, the variation in terms of test accuracy is around 8\%.}
  \label{tab:Acc_Diabetes_Comp}
\end{table*}

\begin{table*}[h!]
  \centering
  \small
  \resizebox{\textwidth}{!}{%
  \begin{tabular}{cc|cccccc|cc}
    \toprule
    &&\multicolumn{3}{c}{\textbf{Best Results}} & \multicolumn{3}{c|}{\textbf{Worst Results}}& & \\
\textit{Layers}&\textit{RU}&\textit{Encoding}&\textit{Bal. Acc. (\%)}&\textit{STD (\%)}&\textit{Encoding}&\textit{Bal. Acc. (\%)}&\textit{STD (\%)}&\textit{$\Delta_{B.Acc.}$(\%)}&\textit{$\Delta_{Bal.Acc.}$ STD (\%)}  \\ \hline 
       2 & False & H-RZ-RY-RX & 66.611111 & 3.737306 & H-RY-RZ-RX & 51.088889 & 1.669129 & 15.522222 & 4.093098 \\
2 & True & H-RZ-RX & 67.700000 & 3.004275 & Amplitude & 56.800000 & 4.229137 & 10.900000 & 5.187607 \\
4 & False & H-RY-RX-RZ & 68.051852 & 4.476554 & RX-RY-RZ & 50.866667 & 1.585466 & 17.185185 & 4.749025 \\
4 & True & H-RZ-RX-RY & 69.616667 & 2.845378 & Amplitude& 60.948148 & 3.178764 & 8.668519 & 4.266230 \\
6 & False & RX-RZ-RY & 69.205556 & 2.348089 & H-RY-RZ-RX & 51.009259 & 1.576511 & 18.196296 & 2.828234 \\
6 & True & H-RY & 69.259259 & 3.192737 & RY-RX-RZ & 63.025926 & 4.719782 & 6.233333 & 5.698238 \\
8 & False & RX-RZ-RY & 69.850000 & 3.590465 & RX-RY-RZ & 51.835185 & 3.097246 & 18.014815 & 4.741769 \\
8 & True & RY & 69.472222 & 2.662523 & Amplitude & 62.650000 & 5.052472 & 6.822222 & 5.711086 \\
10 & False & RX-RZ-RY & 70.222222 & 2.472220 & H-RY-RZ-RX & 50.896296 & 1.281699 & 19.325926 & 2.784713 \\
10 & True & H-RY-RX & 69.512963 & 2.352719 & Amplitude & 62.305556 & 5.110338 & 7.207407 & 5.625908 \\
\bottomrule
\end{tabular}
}
  \caption{Comparison between the best and the worst encoding models in terms of balanced accuracy grouped by the number of parametric layers and use of the re-uploading (RU) technique for classification on the Diabetes dataset. On average, the variation in terms of test accuracy is around 12\%.}
  \label{tab:Acc_Wine_Comp_BalancedAcc}
\end{table*}

\begin{table*}[h!]
  \centering
  \small
  \begin{tabular}{cc|cccccc|cc}
    \toprule
    &&\multicolumn{3}{c}{\textbf{Best Results}} & \multicolumn{3}{c|}{\textbf{Worst Results}}& & \\
\textit{Layers}&\textit{RU}&\textit{Encoding}&\textit{Recall (\%)}&\textit{STD (\%)}&\textit{Encoding}&\textit{Recall (\%)}&\textit{STD (\%)}&\textit{$\Delta_{Rec}$(\%)}&\textit{$\Delta_{Rec}$ STD (\%)}  \\ \hline 
                2 &        False &           H-RY-RX-RZ &              55.000000 &              9.157881 &            H-RY-RZ-RX       &                2.777778 &               4.024199 &              52.222222 &             10.003048 \\
         2 &         True &            H-RZ-RX   &              50.000000 &              5.790637 &            RX-RY-RZ         &               24.629630 &              10.549959 &              25.370370 &             12.034663 \\
         4 &        False &           H-RY-RX-RZ &              53.703704 &              7.759139 &            RX-RY-RZ         &                3.333333 &               4.765496 &              50.370370 &              9.105723 \\
         4 &         True &             H-RZ     &              53.518519 &              7.974678 &      Amplitude              &               41.296296 &               6.593679 &              12.222222 &             10.347565 \\
         6 &        False &           RX-RZ-RY   &              56.111111 &              4.623453 &            H-RY-RZ-RX       &                3.518519 &               5.754331 &              52.592593 &              7.381642 \\
         6 &         True &            RY-RZ     &              54.074074 &              7.943559 &            RX-RZ-RY         &               45.185185 &               9.288021 &               8.888889 &             12.221599 \\
         8 &        False &           RX-RZ-RY   &              55.000000 &              5.923997 &            RX-RY-RZ         &                5.370370 &               8.210109 &              49.629630 &             10.124210 \\
         8 &         True &             X        &              54.629630 &              6.000686 &      Amplitude              &               45.000000 &              11.249143 &               9.629630 &             12.749566 \\
        10 &        False &           H-RY-RX-RZ &              55.740741 &              8.302412 &            H-RY-RZ-RX       &                2.592593 &               3.825169 &              53.148148 &              9.141223 \\
        10 &         True &            RY-RX     &              55.925926 &              2.868877 &            RY-RX-RZ         &               44.814815 &               6.462347 &              11.111111 &              7.070529 \\

\bottomrule
\end{tabular}
  \caption{Comparison between the best and the worst encoding models in terms of recall grouped by the number of parametric layers and use of the re-uploading (RU) technique for classification on the Diabetes dataset. On average, the variation in terms of test recall is around 32\%.}
  \label{tab:Acc_Wine_Comp_Recall}
\end{table*}

\begin{table*}[h!]
  \centering
  \small
  \begin{tabular}{cc|cccccc|cc}
    \toprule
    &&\multicolumn{3}{c}{\textbf{Best Results}} & \multicolumn{3}{c|}{\textbf{Worst Results}}& & \\
\textit{Layers}&\textit{RU}&\textit{Encoding}&\textit{Prec. (\%)}&\textit{STD (\%)}&\textit{Encoding}&\textit{Prec. (\%)}&\textit{STD (\%)}&\textit{$\Delta_{Prec}$(\%)}&\textit{$\Delta_{Prec}$ STD (\%)}  \\ \hline 
               2 &        False &           H-RZ-RX-RY &                 78.315941 &                 8.552728 &      Amplitude &                  48.217717 &                  7.688982 &                 30.098223 &                11.500852 \\
         2 &         True &           H-RZ-RX-RY &                 71.446353 &                 8.296052 &      Amplitude &                  55.572861 &                 11.536338 &                 15.873492 &                14.209560 \\
         4 &        False &           H-RY-RZ-RX &                 73.762838 &                25.861825 &      Amplitude &                  46.398719 &                 12.635906 &                 27.364119 &                28.783678 \\
         4 &         True &            H-RY-RZ &                 68.419916 &                 7.095356 &      Amplitude &                  54.447523 &                  7.264074 &                 13.972393 &                10.154352 \\
         6 &        False &            RY-RX &                 74.367797 &                10.112072 &      Amplitude &                  55.664197 &                  9.544638 &                 18.703599 &                13.905183 \\
         6 &         True &           RX-RY-RZ &                 67.077164 &                 5.693694 &            RY-RX-RZ &                 54.814896 &                  6.863735 &                 12.262268 &                 8.917903 \\
         8 &        False &            RX-RY &                 74.514852 &                 4.434600 &      Amplitude&                  49.007182 &                  5.624648 &                 25.507670 &                 7.162565 \\
         8 &         True &            RY-RX &                 66.513368 &                 5.280583 &            H-RY-RX-RZ &                  54.740224 &                  5.148878 &                 11.773144 &                 7.375330 \\
        10 &        False &           RX-RY-RZ &                 80.769231 &                40.044354 &      Amplitude &                  50.247071 &                 11.931222 &                 30.522160 &                41.784020 \\
        10 &         True &             RY &                 67.616456 &                 6.318357 &      Amplitude &                  53.764180 &                  7.723257 &                 13.852276 &                 9.978494 \\
\bottomrule
\end{tabular}
  \caption{Comparison between the best and the worst encoding models in terms of precision grouped by the number of parametric layers and use of the re-uploading (RU) technique for classification on the Diabetes dataset. On average, the variation in terms of test precision is around 29\%.}
  \label{tab:Acc_Wine_Comp_Precision}
\end{table*}

\begin{table*}[h!]
  \centering
  \small
  \begin{tabular}{cc|cccccc|cc}
    \toprule
    &&\multicolumn{3}{c}{\textbf{Best Results}} & \multicolumn{3}{c|}{\textbf{Worst Results}}& & \\
\textit{Layers}&\textit{RU}&\textit{Encoding}&\textit{F1 (\%)}&\textit{STD (\%)}&\textit{Encoding}&\textit{F1 (\%)}&\textit{STD (\%)}&\textit{$\Delta_{Rec}$(\%)}&\textit{$\Delta_{Rec}$ STD (\%)}  \\ \hline 
               
         2 &        False &           H-RY-RX-RZ        &          55.757982 &          6.366353 &            H-RY-RZ-RX        &           12.543964 &           5.196771 &          43.214017 &          8.218082 \\
         2 &         True &            H-RZ-RX          &          56.406678 &          4.732984 &      Amplitude               &           32.955614 &          10.863263 &          23.451064 &         11.849541 \\
         4 &        False &           H-RY-RX-RZ        &          57.549603 &          6.046461 &            RX-RY-RZ          &           11.462068 &           7.475947 &          46.087536 &          9.615065 \\
         4 &         True &           H-RZ-RX-RY        &          59.314808 &          4.275003 &      Amplitude               &           46.488617 &           4.853552 &          12.826191 &          6.467814 \\
         6 &        False &           RX-RZ-RY          &          59.365833 &          3.365689 &            H-RY-RZ-RX        &           14.478836 &           9.028686 &          44.886996 &          9.635613 \\
         6 &         True &             H-RY            &          58.807912 &          5.260160 &            RX-RZ-RY          &           50.280807 &           7.327488 &           8.527105 &          9.020053 \\
         8 &        False &           RX-RZ-RY          &          59.973117 &          5.146937 &            H-RY-RZ-RX        &           15.768978 &          12.828721 &          44.204139 &         13.822700 \\
         8 &         True &             RY              &          59.438495 &          3.681073 &      Amplitude               &           49.040808 &           8.843486 &          10.397687 &          9.579016 \\
        10 &        False &           RX-RZ-RY          &          60.369697 &          3.583194 &            H-RY-RZ-RX        &            9.261734 &           6.626950 &          51.107964 &          7.533641 \\
        10 &         True &            RY-RX            &          59.772130 &          3.119151 &      Amplitude               &           49.453722 &           7.659514 &          10.318408 &          8.270264 \\
\bottomrule
\end{tabular}
  \caption{Comparison between the best and the worst encoding models in terms of accuracy grouped by the number of parametric layers and use of the re-uploading (RU) technique for classification on the Diabetes dataset. On average, the variation in terms of test F1-score is around 30\%.}
  \label{tab:Acc_Wine_Comp_F1}
\end{table*}

\section{Proposed future methodology}
Given the trajectory analysis and the results obtained, we propose a structured methodology for designing and optimizing Variational Quantum Circuit models. The procedure initiates with a simple implementation and progressively incorporates more sophisticated techniques to improve performance. The first stage involves the application of a \textbf{single rotational gate Angle encoding} strategy. By this, continuous variable features are encoded into the quantum circuit using a single rotational gate (e.g., $RX$, $RY$, or $RZ$) for each feature. This strategy serves as a baseline implementation whose primary goal is to evaluate the model's performance using this simple embedding technique and establish a starting point to enhance the classification performance.

In cases where the single rotational gate strategy yields suboptimal results, more sophisticated Angle encoding approaches are exploited. The idea is to increase progressively the number of rotational gates to encode features along different axes of the qubit state to enhance the model's ability to capture complex patterns in the data, thereby improving classification performance.

This methodology provides a systematic and scalable approach to designing Variational Quantum Circuits models. It guarantees that the complexity of the model does not increase too much, leading to overfit.

\section{Conclusions}
\gls*{qml} is an innovative and rapidly developing field with significant potential for advancement. At this moment in history, due to the limited number of qubits in actual quantum devices and the non-idealities affecting them, it is necessary to consider hybrid solutions in which quantum and classical computing work together. In this regard, new models, such as \gls*{vqc}s are being explored, where the quantum computer implements the \gls*{ml} application and the classical computer reveals useful during the training procedure to update its parameters.\\

This work aims to highlight the impact of quantum embedding on the accuracy of quantum models, with a particular focus on Angle and Amplitude encoding strategies. Specifically, for Angle encoding, a broad and systematic analysis of possible encoding mechanisms is benchmarked. Special attention has been given to the comparability of results, to keep the topology of the variational part constant in every comparison between the adopted encoding strategies. To do so, the number of qubits made available to the models has been kept constant, and PCA has been applied to the considered datasets to reduce the number of features to be processed. This choice has intrinsically disadvantaged the performance of the overall models but guarantees fairness in comparisons. \\

As observed from the results, the accuracy differences between the best and the worst models, considering the same number of layers for the ansatz and the application or non-application of the re-uploading technique, can have an impact of tens of percentage points. Therefore, the encoding strategy should be considered as a hyper-parameter for defining the models, influencing their accuracy performance. Moreover, it is worth noting that no single encoding technique guarantees the best classification, as it depends on the dataset considered. This adds another complexity element in the definition of \gls*{qml} applications, as its development is not strictly related to the formulation of an effective ansatz but opens up the need to investigate the best encoding strategy for a given dataset. \\

In the future, further exploration will be addressed to investigate possible correlations between the encoding techniques and generic input data to identify a strategy to find the best encoding through a-priori analysis on the input dataset, avoiding a time-consuming benchmark procedure today's required. To achieve this, larger datasets should be explored and tested, which has not been possible at the moment due to limited computational resources. Indeed, training and inference of models on larger datasets would take significantly longer. This is also evident from the graphs (\autoref{fig:TimeWine}), where models tested on the Wine dataset (with circuits on 4 qubits) take four times as long as those tested on the Diabetes dataset (3 qubits).
\\ In addition, given the low results in terms of the model's classifications performance, it is considered appropriate to do further investigation on finding the ideal model that is best able to classify by varying all possible hyper-parameters that constitute it. For instance, one could evaluate alternative ansatz topologies constituted by also different entangling gates and the optimal number of layers for each dataset (considering also the application of the reuploading strategy). Once found the optimal hyper-parameters of the model, it could be possible to compare the QML models with the classical ones.\\
Finally, to account for the impact of noise, one could evaluate the classification performance on a noisy simulator by training models in both ideal and noisy simulation environments.

\section*{Data availability statement}
The data that support the findings of this study are openly accessible at the following repository: \url{https://github.com/antotu/VQC-Encoding}
\bibliographystyle{IEEEtran}
\bibliography{Bibliography/biblio}

\end{document}